\documentclass[11pt,a4paper]{article}
\usepackage[hyperref]{emnlp-ijcnlp-2019}
\usepackage{times}
\usepackage{latexsym}
\usepackage{microtype}

\usepackage{url}

\usepackage[T5,T2A,T1]{fontenc}
\usepackage[utf8]{inputenc}

\usepackage{url}
\usepackage{calc}
\usepackage{xcolor}
\usepackage{graphicx}
\usepackage{multirow}
\usepackage{amsthm}
\usepackage{amssymb}
\usepackage{amsmath}
\usepackage{amsfonts}
\usepackage{color, colortbl}
\usepackage{adjustbox}
\usepackage{tabularx}
\usepackage{booktabs}
\usepackage{subcaption}
\usepackage{makecell}
\usepackage{arydshln}

\definecolor{Gray}{gray}{0.92}

\usepackage{bm}

\aclfinalcopy 

\title{Do We Really Need Fully Unsupervised Cross-Lingual Embeddings?}

\author{Ivan Vuli\'c$^{\mathbf{1}}$, ~ Goran Glava\v{s}$^{\mathbf{2}}$, ~ Roi Reichart$^{\mathbf{3}}$, ~ {Anna Korhonen}$^{\mathbf{1}}$\\
$^{\mathbf{1}}$ Language Technology Lab, University of Cambridge \\
$^{\mathbf{2}}$ Data and Web Science Group, University of Mannheim \\
$^{\mathbf{3}}$ Faculty of Industrial Engineering and Management, Technion, IIT\\
\texttt{\{iv250,alk23\}@cam.ac.uk} \\ \texttt{goran@informatik.uni-mannheim.de} \hspace{0.8em} \texttt{roiri@ie.technion.ac.il} }

\date{}

\begin{document}
\maketitle
\begin{abstract}
Recent efforts in cross-lingual word embedding (CLWE) learning have predominantly focused on \textit{fully unsupervised} approaches that project monolingual embeddings into a shared cross-lingual space without any cross-lingual signal. The lack of any supervision makes such approaches conceptually attractive. Yet, their only core difference from (weakly) supervised projection-based CLWE methods is in the way they obtain a \textit{seed dictionary} used to initialize an iterative \textit{self-learning} procedure. The fully unsupervised methods have arguably become more robust, and their \textit{primary use case} is CLWE induction for pairs of resource-poor and distant languages. In this paper, we question the ability of even the most robust unsupervised CLWE approaches to induce meaningful CLWEs in these more challenging settings. A series of bilingual lexicon induction (BLI) experiments with 15 diverse languages (210 language pairs) show that fully unsupervised CLWE methods still fail for a large number of language pairs (e.g., they yield zero BLI performance for 87/210 pairs). Even when they succeed, they never surpass the performance of weakly supervised methods (seeded with 500-1,000 translation pairs) using the same self-learning procedure in any BLI setup, and the gaps are often substantial. These findings call for revisiting the main motivations behind fully unsupervised CLWE methods.
\end{abstract}

\section{Introduction and Motivation}
\label{s:intro}

The wide use and success of monolingual word embeddings in NLP tasks \cite{Turian:2010acl,Chen2014fast} has inspired further research focus on the induction of cross-lingual word embeddings (CLWEs). CLWE methods learn a \textit{shared cross-lingual word vector space} where words with similar meanings obtain similar vectors regardless of their actual language. CLWEs benefit cross-lingual NLP, enabling multilingual modeling of meaning and supporting cross-lingual transfer for downstream tasks and resource-lean languages. CLWEs provide invaluable cross-lingual knowledge for, \textit{inter alia}, bilingual lexicon induction \cite{gouws2015bilbowa,heyman2017bilingual}, information retrieval \cite{vulic2015sigir,litschko2019evaluating}, machine translation \cite{artetxe2018unsupervised,lample2018phrase}, document classification \cite{klementiev2012inducing}, cross-lingual plagiarism detection \cite{glavavs2018resource}, domain adaptation \cite{Ziser2018deep}, cross-lingual POS tagging \cite{gouws2015simple,zhang2016ten}, and cross-lingual dependency parsing \cite{guo2015cross,Sogaard:2015acl}.

The landscape of CLWE methods has recently been dominated by the so-called \textit{projection-based} methods \cite{mikolov2013exploiting,Ruder2018survey,glavas2019howto}. They align two monolingual embedding spaces by learning a projection/mapping based on a training dictionary of translation pairs. Besides their simple conceptual design and competitive performance, their popularity originates from the fact that they rely on rather weak cross-lingual supervision. Originally, the seed dictionaries typically spanned several thousand word pairs \cite{mikolov2013exploiting,faruqui2014improving,xing2015normalized}, but more recent work has shown that CLWEs can be induced with even weaker supervision from small dictionaries spanning several hundred pairs \cite{vulic2016on}, identical strings \cite{smith2017offline}, or even only shared numerals \cite{artetxe2017learning}. 

Taking the idea of reducing cross-lingual supervision to the extreme, the latest CLWE developments almost exclusively focus on \textit{fully unsupervised approaches} \cite[\textit{inter alia}]{conneau2018word,artetxe2018robust,dou2018unsupervised,chen2018unsupervised,alvarez2018gromov,kim2018improving,alaux2019unsupervised,Mohiuddin2019revisiting}: they fully abandon any source of (even weak) supervision and extract the initial seed dictionary by exploiting topological similarities between pre-trained monolingual embedding spaces. Their \textit{modus operandi} can roughly be described by three main components: \textbf{C1)} \textit{unsupervised} extraction of a \textit{seed dictionary}; \textbf{C2)} a \textit{self-learning} procedure that iteratively refines the dictionary to learn  projections of increasingly higher quality; and \textbf{C3)} a set of preprocessing and postprocessing steps (e.g., unit length normalization, mean centering, (de)whitening) \cite{artetxe2018generalizing} that make the entire learning process more robust.

The induction of fully unsupervised CLWEs is an inherently interesting research topic \textit{per se}. Nonetheless, the main practical motivation for developing such approaches in the first place is to facilitate the construction of multilingual NLP tools and widen the access to language technology for resource-poor languages and language pairs. However, the first attempts at fully unsupervised CLWE induction failed exactly for these use cases, as shown by \newcite{sogaard2018on}. Therefore, the follow-up work aimed to improve the robustness of unsupervised CLWE induction by introducing more robust self-learning procedures \cite{artetxe2018robust,Kementchedjhieva:2018conll}. Besides increased robustness, recent work claims that fully unsupervised projection-based CLWEs can even match or surpass their supervised counterparts \cite{conneau2018word,artetxe2018robust,alvarez2018gromov,hoshen2018nonadversarial,Heyman:2019naacl}.



In this paper, we critically examine these claims on robustness and improved performance of unsupervised CLWEs by running a large-scale evaluation in the bilingual lexicon induction (BLI) task on 15 languages (i.e., 210 languages pairs, see~Table~\ref{tab:langs} in \S\ref{s:exp}). The languages were selected to represent different language families and morphological types, as we argue that fully unsupervised CLWEs have been designed to support exactly these setups. However, we show that even the most robust unsupervised CLWE method \cite{artetxe2018robust} still fails for a large number of language pairs: 87/210 BLI setups are unsuccessful, yielding (near-)zero BLI performance. Further, even when the unsupervised method succeeds, it is because the components C2 (self-learning) and C3 (pre-/post-processing) can mitigate the undesired effects of noisy seed lexicon extraction. We then demonstrate that the combination of C2 and C3 with a small provided seed dictionary (e.g., 500 or 1K pairs) outscores the unsupervised method in \textit{all} cases, often with a huge margin, and does not fail for \textit{any} language pair. Furthermore, we show that the most robust unsupervised CLWE approach still fails completely when it relies on monolingual word vectors trained on domain-dissimilar corpora. We also empirically verify that unsupervised approaches cannot outperform weakly supervised approaches also for closely related languages (e.g., Swedish--Danish, Spanish--Catalan).



While the ``no supervision at all'' premise behind fully unsupervised CLWE methods is indeed seductive, our study strongly suggests that future research efforts should revisit the main motivation behind these methods and focus on designing even more robust solutions, given their current inability to support a wide spectrum of language pairs. In hope of boosting induction of CLWEs for more diverse and distant language pairs, we make all 210 training and test dictionaries used in this work publicly available at: \url{https://github.com/ivulic/panlex-bli}.

\section{Methodology}
\label{s:methods}
We now dissect a general framework for unsupervised CLWE learning, and show that the ``bag of tricks of the trade'' used to increase their robustness (which often slips under the radar) can be equally applied to (weakly) supervised projection-based approaches, leading to their fair(er) comparison.


\subsection{Projection-Based CLWE Approaches}
In short, projection-based CLWE methods learn to (linearly) align independently trained monolingual spaces $\bm{X}$ and $\bm{Z}$, using a word translation dictionary $D_0$ to guide the alignment process. Let $\bm{X}_D \subset \bm{X}$ and $\bm{Z}_D \subset \bm{Z}$ be the row-aligned subsets of monolingual spaces containing vectors of aligned words from $D_0$. Alignment matrices $\bm{X}_D$ and $\bm{Z}_D$ are then used to learn orthogonal transformations $\bm{W}_x$ and $\bm{W}_z$ that define the joint bilingual space $\bm{Y} = \bm{XW}_x \cup \bm{ZW}_z$. While supervised projection-based CLWE models learn the mapping using a provided external (clean) dictionary $D_0$, their unsupervised counterparts automatically induce the seed dictionary in an unsupervised way (C1) and then refine it in an iterative fashion (C2).            

\begin{figure}
    \centering
    \includegraphics[width=0.99\linewidth]{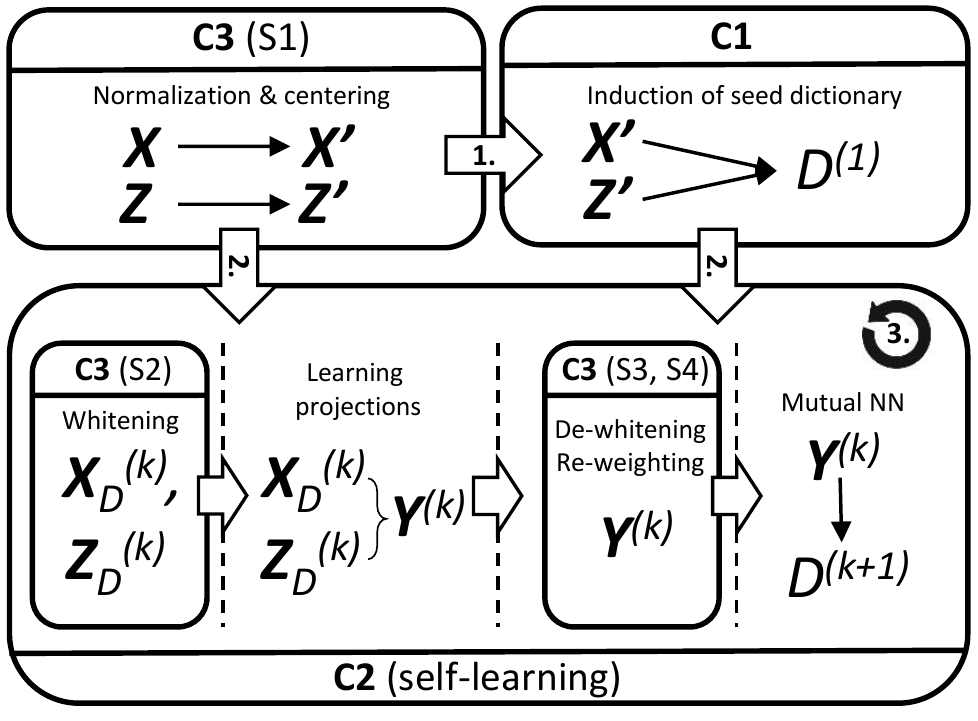}
    \vspace{-3mm}
    \caption{General unsupervised CLWE approach.}
    \label{fig:unsup_clwe}
    \vspace{-2.5mm}
\end{figure}

\vspace{1.8mm}
\noindent \textbf{Unsupervised CLWEs.} These methods first induce a seed dictionary $D^{(1)}$ leveraging only two unaligned monolingual spaces (C1). While the algorithms for unsupervised seed dictionary induction differ, they all strongly rely on the assumption of similar topological structure between the two pretrained monolingual spaces. Once the seed dictionary is obtained, the two-step iterative self-learning procedure (C2) takes place: 1) a dictionary $D^{(k)}$ is first used to learn the joint space $\bm{Y}^{(k)} = \bm{X{W}}^{(k)}_x \cup \bm{Z{W}}^{(k)}_z$; 2) the nearest neighbours in $\bm{Y}^{(k)}$ then form the new dictionary $D^{(k+1)}$. We illustrate the general structure in Figure \ref{fig:unsup_clwe}. 



A recent empirical survey paper \cite{glavas2019howto} has compared a variety of latest unsupervised CLWE methods \cite{conneau2018word,alvarez2018gromov,hoshen2018nonadversarial,artetxe2018robust} in several downstream tasks (e.g., BLI, cross-lingual information retrieval, document classification). The results of their study indicate that the \textsc{vecmap} model of \newcite{artetxe2018robust} is by far the most robust and best performing unsupervised CLWE model. For the actual results and analyses, we refer the interested reader to the original paper of \newcite{glavas2019howto}. Another recent evaluation paper \cite{doval2019onthe} as well as our own preliminary BLI tests (not shown for brevity) have further verified their findings. We thus focus on \textsc{vecmap} in our analyses, and base the following description of the components C1-C3 on that model.

%



\subsection{Three Key Components}

\noindent \textbf{C1. Seed Lexicon Extraction.} \textsc{vecmap} induces the initial seed dictionary using the following heuristic: monolingual similarity distributions for words with similar meaning will be similar across languages.\footnote{For instance, \textit{zwei} and \textit{two} will have similar distributions of similarities over their respective language vocabularies -- \textit{zwei} is expected to be roughly as (dis)similar to \textit{drei} and \textit{Katze} as \textit{two} is to \textit{three} and \textit{cat}.} The monolingual similarity distributions for the two languages are given as rows (or columns; the matrices are symmetric) of $\bm{M}_x = \bm{X}\bm{X}^T$ and $\bm{M}_z = \bm{Z}\bm{Z}^T$. For the distributions of similarity scores to be comparable, the values in each row of $\bm{M}_x$ and $\bm{M}_z$ are first sorted. The initial dictionary $D^{(1)}$ is finally obtained by searching for mutual nearest neighbours between the rows of $\sqrt{\bm{M}_x}$ and of $\sqrt{\bm{M}_z}$.         

\vspace{1.8mm}
\noindent \textbf{C2. Self-Learning.} Not counting the preprocessing and postprocessing steps (component C3), self-learning then \textit{iteratively} repeats two steps: 

\vspace{1.4mm}

\noindent \textbf{1)} Let $\bm{D}^{(k)}$ be the binary matrix indicating the aligned words in the dictionary $D^{(k)}$.\footnote{I.e., $\bm{D}^{(k)}_{ij} = 1 \iff$ the $i$-th word of one language and the $j$-th word of the other are a translation pair in $D^{(k)}$.} The orthogonal transformation matrices are then obtained as $\bm{W}^{(k)}_x = \bm{U}$ and $\bm{W}^{(k)}_z = \bm{V}$, where $\bm{U}\bm{\Sigma}\bm{V}^T$ is the singular value decomposition of the matrix $\bm{X}^T\bm{D}^{(k)}\bm{Z}$. The cross-lingual space of the $k$-th iteration is then $\bm{Y}^{(k)} = \bm{X}\bm{W}^{(k)}_x \cup \bm{Z}\bm{W}^{(k)}_z$.   

\vspace{1.4mm}

\noindent \textbf{2)} The new dictionary $D^{(k+1)}$ is then built by identifying nearest neighbours in $\bm{Y}^{(k)}$. These can be easily extracted from the matrix $\bm{P} = \bm{X}\bm{W}^{(k)}_x( \bm{Z}\bm{W}^{(k)}_z)^T$. \textit{All} nearest neighbours can be used, or additional \textit{symmetry} constraints can be imposed to extract only mutual nearest neighbours: all pairs of indices ($i, j$) for which $\bm{P}_{ij}$ is the largest value both in row $i$ and column $j$.
\vspace{1.4mm}


The above procedure, however, often converges to poor local optima. To remedy for this, the second step (i.e., dictionary induction) is extended with techniques that make self-learning more robust. First, the vocabularies of $\bm{X}$ and $\bm{Z}$ are cut to the top $k$ most frequent words.\footnote{This is done to prevent spurious nearest neighbours consisting of infrequent words with unreliable vectors.} Second, similarity scores in $\bm{P}$ are kept with  probability $p$, and set to zero otherwise. This \textit{dropout} allows for a wider exploration of possible word pairs in the dictionary and contributes to escaping poor local optima given the noisy seed lexicon in the first iterations. 


\vspace{1.8mm}
\noindent \textbf{C3. Preprocessing and Postprocessing Steps.} While iteratively learning orthogonal transformations $\bm{W}_{x}$ and $\bm{W}_{z}$ for $\bm{X}$ and $\bm{Z}$ is the central step of unsupervised projection-based CLWE methods, preprocessing and postprocessing techniques are additionally applied before and after the transformation. While such techniques are often overlooked in model comparisons, they may have a great impact on the model's final performance, as we validate in \S\ref{s:results}. 
%
We briefly summarize two pre-processing (S1 and S2) and post-processing (S3 and S4) steps used in our evaluation, originating from the framework of \newcite{artetxe2018generalizing}.

\vspace{1.4mm}
\noindent S1) \textit{Normalization and mean centering}. We first apply unit length normalization: all vectors in $\bm{X}$ and $\bm{Z}$ are normalized to have a unit Euclidean norm. Following that, $\bm{X}$ and $\bm{Z}$ are mean centered dimension-wise and then again length-normalized.

\vspace{1.4mm}
\noindent S2) \textit{Whitening}. ZCA whitening~\citep{Bell1997} is applied on (S1-processed) $\bm{X}$ and $\bm{Z}$: it transforms the matrices such that each dimension has unit variance and that the dimensions are uncorrelated. Intuitively, the vector spaces are easier to align along directions of high variance.

\vspace{1.4mm}
\noindent S3) \textit{Dewhitening}. A transformation inverse to S2: for improved performance it is important to restore the variance information after the projection, if whitening was applied in S2~\citep{artetxe2018generalizing}.

\vspace{1.2mm}

\noindent S4) \textit{Symmetric re-weighting}. This step attempts to further align the embeddings in the cross-lingual embedding space by measuring how well a dimension in the space correlates across languages for the current iteration dictionary $D^{(k)}$.\footnote{More formally, assume that we are working with matrices $\bm{X}$ and $\bm{Z}$ that already underwent all transformations described in S1-S3. Another matrix $\bm{D}$ represents the current bilingual dictionary $D$: $D_{ij} = 1$ if the $i^{th}$ source word is translated by the $j^{th}$ target word and $D_{ij}=0$ otherwise. Then, given the singular value decomposition $\bm{USV^T} = \bm{X^TDZ}$, the final re-weighted projection matrices are $\bm{W}_x=\bm{US}^{\frac{1}{2}}$ (and $\bm{W}_z=\bm{VS}^{\frac{1}{2}}$. We refer the reader to \cite{artetxe2018generalizing} and \cite{artetxe2018robust} for more details.} 
The best results are obtained when re-weighting is neutral to the projection direction, that is, when it is applied symmetrically in both languages.

\vspace{1.4mm}
In the actual implementation S1 is applied only once, before self-learning. S2, S3 and S4 are applied in each self-learning iteration. 


\vspace{1.8mm}
\noindent \textbf{Model Configurations.}
Note that C2 and C3 can be equally used on top of any (provided) seed lexicon (i.e., $D^{(1)}$:=$D_0$) to enable weakly supervised learning, as we propose here. In fact, the variations of the three key components, C1) seed lexicon, C2) self-learning, and C3) preprocessing and postprocessing, construct various model configurations which can be analyzed to probe the importance of each component in the CLWE induction process. A selection of representative configurations evaluated later in \S\ref{s:results} is summarized in Table~\ref{tab:config}.

\begin{table*}[t!]
\def\arraystretch{0.97}
\centering
{\small
\begin{adjustbox}{max width=\linewidth}
\begin{tabular}{l lll}
\toprule
{\bf Configuration} & {C1} & {C2} & {C3} \\
\cmidrule(lr){2-4}
\textsc{unsupervised} & {unsupervised} & {all tested, we always report the best one} & {S1-S4 (\textsc{full})} \\
\hdashline
\rowcolor{Gray}
\textsc{orthg-super} & {provided} & {--} & {length normalization only (partial S1)} \\
\textsc{orthg+sl+sym} & {provided} & {symmetric: mutual nearest neighbours} & {length normalization only (partial S1)} \\
\hdashline
\rowcolor{Gray}
\textsc{full-super} & {provided} & {--} & {S1-S4 (\textsc{full})} \\
\textsc{full+sl} & {provided} & {\cite{artetxe2018robust} with dropout} & {S1-S4 (\textsc{full})} \\
\rowcolor{Gray}
\textsc{full+sl+nod} & {provided} & {\cite{artetxe2018robust} w/o dropout} & {S1-S4 (\textsc{full})} \\
\textsc{full+sl+sym} & {provided} & {symmetric: mutual nearest neighbours, w/o dropout} & {S1-S4 (\textsc{full})} \\
\bottomrule
\end{tabular}
\end{adjustbox}
}
\vspace{-2mm}
\caption{Configurations obtained by varying components C1, C2, and C3 used in our empirical comparison in \S\ref{s:results}.}
\label{tab:config}
\vspace{-1.5mm}
\end{table*}
\begin{table}[t]
\def\arraystretch{0.97}
\centering
{\footnotesize
\begin{tabularx}{\linewidth}{l ll l}
\toprule 
{\bf Language} & {Family} & {Type} & {ISO 639-1} \\
\cmidrule(lr){2-4}
{Bulgarian} & {IE: Slavic} & {fusional} & {\textsc{bg}} \\
{Catalan} & {IE: Romance} & {fusional} & {\textsc{ca}} \\
{Esperanto} & {-- (constructed)} & {agglutinative} & {\textsc{eo}} \\
{Estonian} & {Uralic} & {agglutinative} & {\textsc{et}} \\
{Basque} & {-- (isolate)} & {agglutinative} & {\textsc{eu}} \\
{Finnish} & {Uralic} & {agglutinative} & {\textsc{fi}} \\
{Hebrew} & {Afro-Asiatic} & {introflexive} & {\textsc{he}} \\
{Hungarian} & {Uralic} & {agglutinative} & {\textsc{hu}} \\
{Indonesian} & {Austronesian} & {isolating} & {\textsc{id}} \\
{Georgian} & {Kartvelian} & {agglutinative} & {\textsc{ka}} \\
{Korean} & {Koreanic} & {agglutinative} & {\textsc{ko}} \\
{Lithuanian} & {IE: Baltic} & {fusional} & {\textsc{lt}} \\
{Bokm\aa l} & {IE: Germanic} & {fusional} & {\textsc{no}} \\
{Thai} & {Kra-Dai} & {isolating} & {\textsc{th}} \\
{Turkish} & {Turkic} & {agglutinative} & {\textsc{tr}} \\
\bottomrule
\end{tabularx}
}
\vspace{-1.5mm}
\caption{The list of 15 languages from our main BLI experiments along with their corresponding language family (IE = Indo-European), broad morphological type, and their ISO 639-1 code.}
\label{tab:langs}
\vspace{-2.5mm}
\end{table}

\section{Experimental Setup}
\label{s:exp}

\noindent \textbf{Evaluation Task.}
Our task is \textit{bilingual lexicon induction} (BLI). It has become the \textit{de facto} standard evaluation for projection-based CLWEs \cite{Ruder2018survey,glavas2019howto}. In short, after a shared CLWE space has been induced, the task is to retrieve target language translations for a test set of source language words. Its lightweight nature allows us to conduct a comprehensive evaluation across a large number of language pairs.\footnote{While BLI is an intrinsic task, as discussed by \newcite{glavas2019howto} it is a strong indicator of CLWE quality also for downstream tasks: relative performance in the BLI task correlates well with performance in cross-lingual information retrieval \cite{litschko2018unsupervised} or natural language inference \cite{conneau2018xnli}. More importantly, it also provides a means to analyze whether a CLWE method manages to learn anything meaningful at all, and can indicate ``unsuccessful'' CLWE induction (e.g., when BLI performance is similar to a random baseline): detecting such CLWEs is especially important when dealing with fully unsupervised models.} Since BLI is cast as a ranking task, following \newcite{glavas2019howto} we use mean average precision (MAP) as the main evaluation metric: in our BLI setup with only one correct translation for each ``query'' word, MAP is equal to mean reciprocal rank (MRR).\footnote{MRR is more informative than the more common \textit{Precision@1 (P@1)}; our main findings are also valid when P@1 is used; we do not report the results for brevity.}


\vspace{1.8mm}
\noindent \textbf{(Selection of) Language Pairs.}
Our selection of test languages is guided by the following goals: \textbf{a)} following recent initiatives in other NLP research (e.g., for language modeling) \cite{Cotterell:2018naacl,Gerz2018on}, we aim to ensure the coverage of different genealogical and typological language properties, and \textbf{b)} we aim to analyze a large set of language pairs and offer new evaluation data which extends and surpasses other work in the CLWE literature. These two properties will facilitate analyses between (dis)similar language pairs and offer a comprehensive set of evaluation setups that test the robustness and portability of fully unsupervised CLWEs. The final list of 15 diverse test languages is provided in Table~\ref{tab:langs}, and includes samples from different languages types and families. We run BLI evaluations for all language pairs in both directions, for a total of 15$\times$14=210 BLI setups.


\vspace{1.8mm}
\noindent \textbf{Monolingual Embeddings.}
We use the 300-dim vectors of \newcite{Grave:2018lrec} for all 15 languages, pretrained on Common Crawl and Wikipedia with fastText \cite{Bojanowski:2017tacl}.\footnote{Experiments with other monolingual vectors such as the original fastText and skip-gram \cite{Mikolov2013distributed} trained on Wikipedia show the same trends in the final results.} We trim all vocabularies to the 200K most frequent words.


\vspace{1.8mm}
\noindent \textbf{Training and Test Dictionaries.} They are derived from PanLex \cite{Baldwin2010panlex,Kamholz2014panlex}, which was used in prior work on cross-lingual word embeddings \cite{Duong:2016emnlp,Vulic:2017emnlp}. PanLex currently spans around 1,300 language varieties with over 12M expressions: it offers some support and supervision also for low-resource language pairs \cite{Adams:2017eacl}. For each source language ($L_1$), we automatically translate their vocabulary words (if they are present in PanLex) to all 14 target ($L_2$) languages. To ensure the reliability of the translation pairs, we retain only unigrams found in the vocabularies of the respective $L_2$ monolingual spaces which scored above a PanLex-predefined threshold. 

\begin{table*}[t]
\def\arraystretch{0.97}
\centering
{\small
\begin{tabularx}{\linewidth}{l XXXXXXXX}
\toprule
{} & {\textsc{bg}-*} & {\textsc{ca}-*} & {\textsc{eo}-*} & {\textsc{et}-*} & {\textsc{eu}-*} & {\textsc{fi}-*} & {\textsc{he}-*} & {\textsc{hu}-*} \\
\cmidrule(lr){2-2} \cmidrule(lr){3-3} \cmidrule(lr){4-4} \cmidrule(lr){5-5} \cmidrule(lr){6-6} \cmidrule(lr){7-7} \cmidrule(lr){8-8} \cmidrule(lr){9-9} 
\textsc{unsupervised} & {0.208} & {0.224} & {0.128} & {0.155} & {0.036} & {0.181} & {0.186} & {0.206} \\
\textit{Unsuccessful setups} & {\it 3/14} & {\it 2/14} & {\it 3/14} & {\it 6/14} & {\it 10/14} & {\it 4/14} & {\it 2/14} & {\it 3/14} \\
\hdashline
\rowcolor{Gray}
\textsc{5k:orthg-super} & {0.258} & {0.237} & {0.201} & {0.210} & {0.151} & {0.233} & {0.198} & {0.259} \\
\textsc{5k:orthg+sl+sym} & {0.281} & {0.264} & {0.219} & {0.225} & {0.164} & {0.256} & {0.217} & {0.283} \\
\rowcolor{Gray}
\textsc{5k:full-super} & \underline{0.343} & \underline{0.335} & \underline{0.304} & \underline{0.301} & \underline{0.228} & \underline{0.324} & \underline{0.287} & \underline{0.354} \\
\textsc{5k:full+sl}  & {0.271} & {0.262} & {0.240} & {0.236} & {0.161} & {0.260} & {0.217} & {0.282} \\
\rowcolor{Gray}
\textsc{5k:full+sl+nod} & {0.316} & {0.311} & {0.295} & {0.276} & {0.204} & {0.320} & {0.260} & {0.330} \\
\textsc{5k:full+sl+sym} & {\bf 0.361} & {\bf 0.356} & {\bf 0.336} & {\bf 0.316} & {\bf 0.244} & {\bf 0.348} & {\bf 0.294} & {\bf 0.374}\\
\hdashline
\rowcolor{Gray}
\textsc{1k:orthg-super} & {0.104} & {0.088} & {0.065} & {0.082} & {0.049} & {0.088} & {0.066} & {0.101} \\
\textsc{1k:orthg+sl+sym}  & {0.203} & {0.167} & {0.106} & {0.157} & {0.079} & {0.168} & {0.133} & {0.191} \\
\rowcolor{Gray}
\textsc{1k:full-super} & {0.146} & {0.129} & {0.098} & {0.117} & {0.065} & {0.117} & {0.096} & {0.143} \\
\textsc{1k:full+sl} & {0.268} & {0.260} & {0.238} & {0.232} & {0.158} & {0.257} & {0.217} & {0.279} \\
\rowcolor{Gray}
\textsc{1k:full+sl+nod}  & \underline{0.312} & \underline{0.307} & \underline{0.284} & \underline{0.272} & \underline{0.197} & \underline{0.311} & \underline{0.255} & \underline{0.327} \\
\textsc{1k:full+sl+sym} & {\bf 0.341} & {\bf 0.327} & {\bf 0.302} & {\bf 0.293} & {\bf 0.212} & {\bf 0.329} & {\bf 0.268} & {\bf 0.354} \\
\bottomrule
\end{tabularx}
\smallskip \smallskip
\begin{tabularx}{\linewidth}{l XXXXXXXX}
\toprule
{} & {\textsc{id}-*} & {\textsc{ka}-*} & {\textsc{ko}-*} & {\textsc{lt}-*} & {\textsc{no}-*} & {\textsc{th}-*} & {\textsc{tr}-*} & {\textbf{Avg}} \\
\cmidrule(lr){2-2} \cmidrule(lr){3-3} \cmidrule(lr){4-4} \cmidrule(lr){5-5} \cmidrule(lr){6-6} \cmidrule(lr){7-7} \cmidrule(lr){8-8} \cmidrule(lr){9-9} 
\textsc{unsupervised} & {0.110} & {0.106} & {0.001} & {0.179} & {0.239} & {0.000} & {0.133} & {0.140} \\
\textit{Unsuccessful setups} & {\it 7/14} & {\it 6/14} & {\it 14/14} & {\it 4/14} & {\it 3/14} & {\it 14/14} & {\it 6/14} & {\it 87/210} \\
\hdashline
\rowcolor{Gray}
\textsc{5k:orthg-super} & {0.199} & {0.163} & {0.154} & {0.194} & {0.250} & {0.109} & {0.207} & {0.201} \\
\textsc{5k:orthg+sl+sym} & {0.216} & {0.177} & {0.166} & {0.212} & {0.273} & {0.117} & {0.226} & {0.220} \\
\rowcolor{Gray}
\textsc{5k:full-super} & \underline{0.261} & \underline{0.250} & \underline{0.239} & \underline{0.302} & \underline{0.332} & {\bf 0.154} & \underline{0.283} & {0.286} \\
\textsc{5k:full+sl}  & {0.180} & {0.191} & {0.152} & {0.217} & {0.274} & {0.056} & {0.204} & {0.214} \\
\rowcolor{Gray}
\textsc{5k:full+sl+nod} & {0.220} & {0.229} & {0.207} & {0.272} & {0.318} & {0.106} & {0.253} & {0.261} \\
\textsc{5k:full+sl+sym} & {\bf 0.272} & {\bf 0.263} & {\bf 0.251} & {\bf 0.310} & {\bf 0.356} & \underline{0.148} & {\bf 0.299} & {\bf 0.302}\\
\hdashline
\rowcolor{Gray}
\textsc{1k:orthg-super} & {0.069} & {0.050} & {0.040} & {0.067} & {0.099} & {0.034} & {0.068} & {0.071} \\
\textsc{1k:orthg+sl+sym}  & {0.119} & {0.092} & {0.063} & {0.135} & {0.186} & {0.052} & {0.129} & {0.132} \\
\rowcolor{Gray}
\textsc{1k:full-super} & {0.089} & {0.079} & {0.061} & {0.111} & {0.127} & {0.044} & {0.091} & {0.101} \\
\textsc{1k:full+sl} & {0.180} & {0.185} & {0.148} & {0.220} & {0.274} & {0.054} & {0.204} & {0.212} \\
\rowcolor{Gray}
\textsc{1k:full+sl+nod}  & \underline{0.216} & \underline{0.223} & \underline{0.197} & \underline{0.269} & \underline{0.315} & \underline{0.096} & \underline{0.248} & \underline{0.255} \\
\textsc{1k:full+sl+sym} & {\bf 0.243} & {\bf 0.237} & {\bf 0.203} & {\bf 0.284} & {\bf 0.337} & {\bf 0.103} & {\bf 0.274} & {\bf 0.274} \\
\bottomrule
\end{tabularx}
}
\vspace{-1.5mm}
\caption{BLI scores (MRR) for all model configurations. The scores are averaged over all experimental setups where each of the 15 languages is used as $L_1$: e.g., \textsc{ca}-* means that the translation direction is from Catalan (\textsc{ca}) as source ($L_1$) to each of the remaining 14 languages listed in Table~\ref{tab:langs} as targets ($L_2$), and we average over the corresponding 14 \textsc{ca}-* BLI setups. $5k$ and $1k$ denote the seed dictionary size for (weakly) supervised methods ($D_0$). \textit{Unsuccessful setups} refer to the number of BLI experimental setups with the fully \textsc{unsupervised} model that yield an MRR score $\leq0.01$. The \textbf{Avg} column refers to the averaged MRR scores of each model configuration over all 15$\times$14=210 BLI setups. The highest scores for two different seed dictionary sizes in each column are in bold, the second best are underlined. See Table~\ref{tab:config} for the brief description of all model configurations in the comparison. Full results for each particular language pair are available in the supplemental material.}
\label{tab:avg-per-lang}
\vspace{-1mm}
\end{table*}

\begin{table}[!t]
\def\arraystretch{0.99}
\centering
{\small
\begin{tabularx}{\linewidth}{l lX lX}
\toprule
{} & \multicolumn{2}{l}{$|D_{0}|=5k$} & \multicolumn{2}{l}{$|D_{0}|=1k$} \\
\cmidrule(lr){2-3} \cmidrule(lr){4-5}
{} & {\bf Unsuc.} & {\bf Win} & {\bf Unsuc.} & {\bf Win} \\
\cmidrule(lr){2-2} \cmidrule(lr){3-3}  \cmidrule(lr){4-4} \cmidrule(lr){5-5} 
\textsc{unsupervised} & {87 (94)} & {0} & {87 (94)} & {0} \\
\hdashline
\rowcolor{Gray}
\textsc{orthg-super} & {0 (2)} & {0} & {2 (82)} & {0} \\
\textsc{orthg+sl+sym} & {0 (1)} & {0} & {1 (34)} & {0} \\
\rowcolor{Gray}
\textsc{full-super} & {0 (0)} & {46} & {0 (41)} & {0} \\
\textsc{full+sl} & {0 (7)} & {0} & {0 (9)} & {0} \\
\rowcolor{Gray}
\textsc{full+sl+nod} & {0 (1)} & {7} & {0 (3)} & {33} \\
\textsc{full+sl+sym} & {0 (0)} & {157} & {0 (0)} & {177} \\
\bottomrule
\end{tabularx}
}
\vspace{-1.5mm}
\caption{Summary statistics computed over all 15$\times$14=210 BLI setups. \textbf{a) Unsuc.} denotes the total number of unsuccessful setups, where a setup is considered unsuccessful if MRR $\leq$ 0.01 or MRR $\leq$ 0.05 (in the parentheses); \textbf{b) Win} refers to the total number of ``winning'' setups, that is, for all language pairs it counts how many times each particular model yields the best overall MRR score. We compute separate statistics for two settings ($|D_0|=1k$ and $|D_0|=5k$).}
\label{tab:avg-runs}
\vspace{-1.5mm}
\end{table}

As in prior work \cite{conneau2018word,glavas2019howto}, we then reserve the 5K pairs created from the more frequent $L_1$ words for training, while the next 2K pairs are used for test. Smaller training dictionaries (1K and 500 pairs) are created by again selecting pairs comprising the most frequent $L_1$ words. 


\vspace{1.8mm}
\noindent \textbf{Training Setup.}
In all experiments, we set the hyper-parameters to values that were tuned in prior research. When extracting the \textsc{unsupervised} seed lexicon, the 4K most frequent words of each language are used; self-learning operates on the 20K most frequent words of each language; with dropout the keep probability $p$ is 0.1; CSLS with $k=10$ nearest neighbors \cite{artetxe2018robust}. 


Again, Table~\ref{tab:config} lists the main model configurations in our comparison. For the fully \textsc{unsupervised} model we always report the best performing configuration after probing different self-learning strategies (i.e., \textsc{+sl}, \textsc{+sl+nod}, and \textsc{+sl+sym} are tested). The results for \textsc{unsupervised} are always reported as averages over 5 restarts: this means that with \textsc{unsupervised} we count BLI setups as unsuccessful only if the results are close to zero in all 5/5 runs. \textsc{orthg-super} is the standard supervised model with orthogonal projections from prior work \cite{smith2017offline,glavas2019howto}.

\section{Results and Discussion}
\label{s:results}
Main BLI results averaged over each source language ($L_1$) are provided in Table~\ref{tab:avg-per-lang} and Table~\ref{tab:avg-runs}. We now summarize and discuss the main findings across several dimensions of comparison.

\begin{figure}[t]
    \centering
    \includegraphics[width=0.99\linewidth]{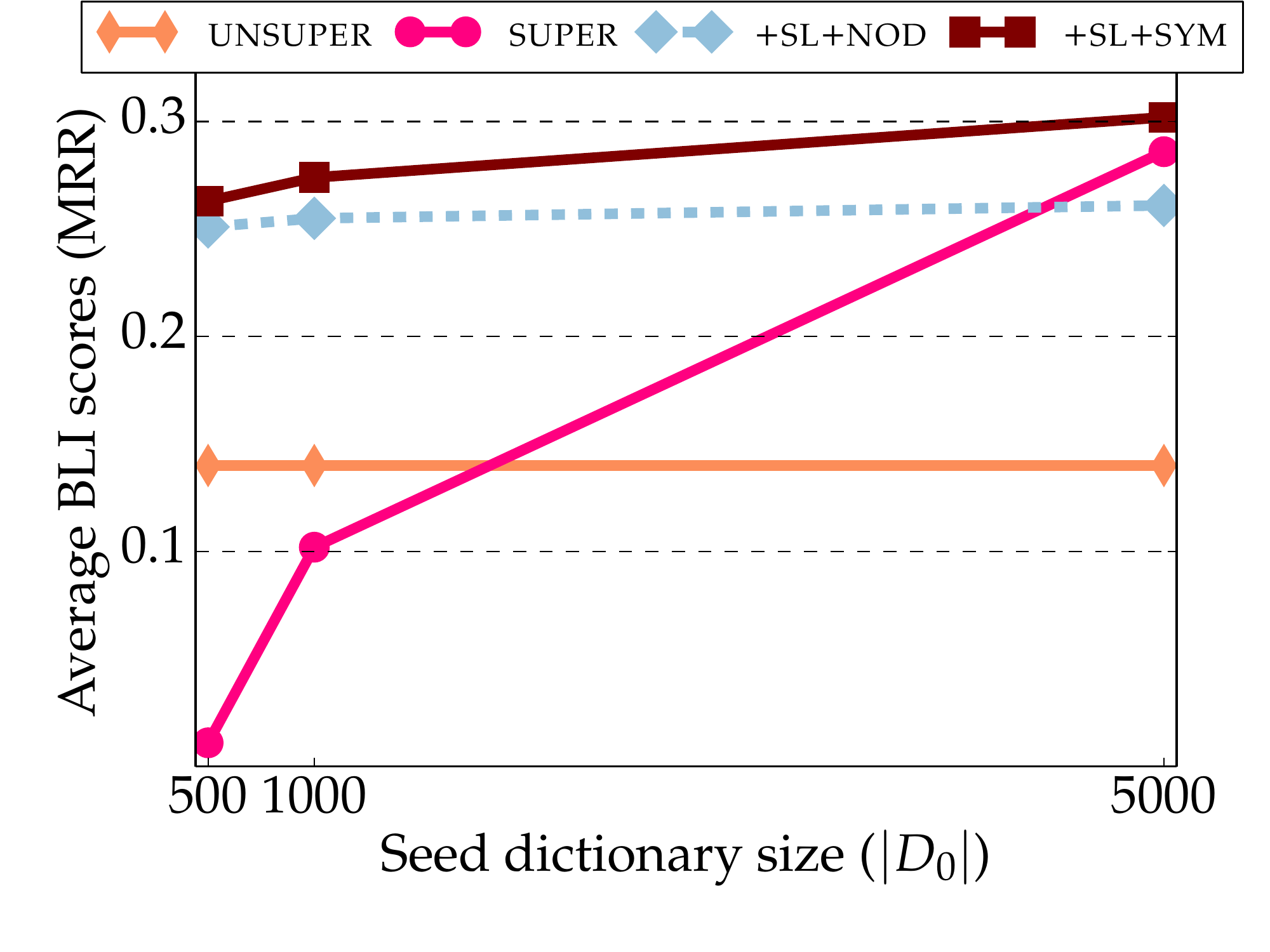}
    \vspace{-4.5mm}
    \caption{A comparison of average BLI scores with different seed dictionary sizes $D_0$ between a fully unsupervised method (\textsc{unsuper}), a supervised method without self-learning (\textsc{super}), and two best performing weakly supervised methods with self learning (\textsc{+sl+nod} and \textsc{+sl+sym}). While \textsc{super} without self-learning displays a steep drop in performance with smaller seed dictionaries, there is only a slight decrease when self-learning is used: e.g., 500 translation pairs are still sufficient to initialize robust self-learning.}
    \vspace{-1.5mm}
\label{fig:d0}
\end{figure}

\vspace{1.8mm}
\noindent \textbf{Unsupervised vs. (Weakly) Supervised.} First, when exactly the same components C2 and C3 are used, \textsc{unsupervised} is unable to outperform a (weakly) supervised \textsc{full+sl+sym} variant, and the gap in final performance is often substantial. In fact, \textsc{full+sl+sym} and \textsc{full+sl+nod} outperform the best \textsc{unsupervised} for all 210/210 BLI setups: we observe the same phenomenon with varying dictionary sizes, that is, it equally holds when we seed self-learning with 5K, 1K, and 500 translation pairs, see also Figure~\ref{fig:d0}. This also suggests that the main reason why \textsc{unsupervised} approaches were considered on-par with supervised approaches in prior work \cite{conneau2018word,artetxe2018robust} is because they were not compared under fair circumstances: while \textsc{unsupervised} relied heavily on the components C2 and C3, these were omitted when running supervised baselines. Our unbiased comparison reveals that there is a huge gap even when supervised projection-based approaches consume only several hundred translation pairs to initiate self-learning.


\vspace{1.8mm}
\noindent \textbf{Are Unsupervised CLWEs Robust?}
The results also indicate that, contrary to the beliefs established by very recent work \cite{artetxe2018robust,Mohiuddin2019revisiting}, fully \textsc{unsupervised} approaches are still prone to getting stuck in local optima, and still suffer from robustness issues when dealing with distant language pairs: 87 out of 210 BLI setups ($=41.4\%$) result in (near-)zero BLI performance, see also Table~\ref{tab:avg-runs}. At the same time, weakly supervised methods with a seed lexicon of 1k or 500 pairs do not suffer from the robustness problem and always converge to a good solution, as also illustrated by the results reported in Table~\ref{tab:unsuc-pairs}.


\begin{table*}[t]
\def\arraystretch{0.99}
\centering
{\small
\begin{adjustbox}{max width=\linewidth}
\begin{tabular}{l cccccccccc}
\toprule
{} & {\textsc{bg-eu}} & {\textsc{eu-tr}} & {\textsc{fi-ko}} & {\textsc{id-et}} & {\textsc{id-th}} & {\textsc{ka-fi}} & {\textsc{ka-id}} & {\textsc{ko-tr}} &  {\textsc{no-eu}} & {\textsc{tr-th}} \\
\cmidrule(lr){2-2} \cmidrule(lr){3-3} \cmidrule(lr){4-4} \cmidrule(lr){5-5} \cmidrule(lr){6-6} \cmidrule(lr){7-7} \cmidrule(lr){8-8} \cmidrule(lr){9-9}  \cmidrule(lr){10-10}  \cmidrule(lr){11-11} 
\textsc{unsupervised} & {0.005} & {0.000} & {0.000} & {0.000} & {0.000} & {0.004} & {0.000} & {0.000} & {0.000} & {0.000} \\
\hdashline
\textsc{1k:full+sl+sym} & {\bf 0.279} & {\bf 0.212} & {\bf 0.211} & {\bf 0.213} & {\bf 0.226} & {\bf 0.306} & {\bf 0.155} & {\bf 0.279} & {\bf 0.300} & {\bf 0.137} \\
\textsc{500:full+sl+sym} & {0.245} & {0.189} & {0.192} & {0.195} & {0.188} & {0.285} & {0.138} & {0.264} & {0.266} & {0.109} \\
\bottomrule
\end{tabular}
\end{adjustbox}
}
\vspace{-1.5mm}
\caption{Results for a selection of BLI setups which were unsuccessful with the \textsc{unsupervied} CLWE method.}
\label{tab:unsuc-pairs}
\vspace{-1.5mm}
\end{table*}

\begin{figure*}[!t]
    \centering
    \begin{subfigure}[t]{0.328\linewidth}
        \centering
        \includegraphics[width=0.98\linewidth]{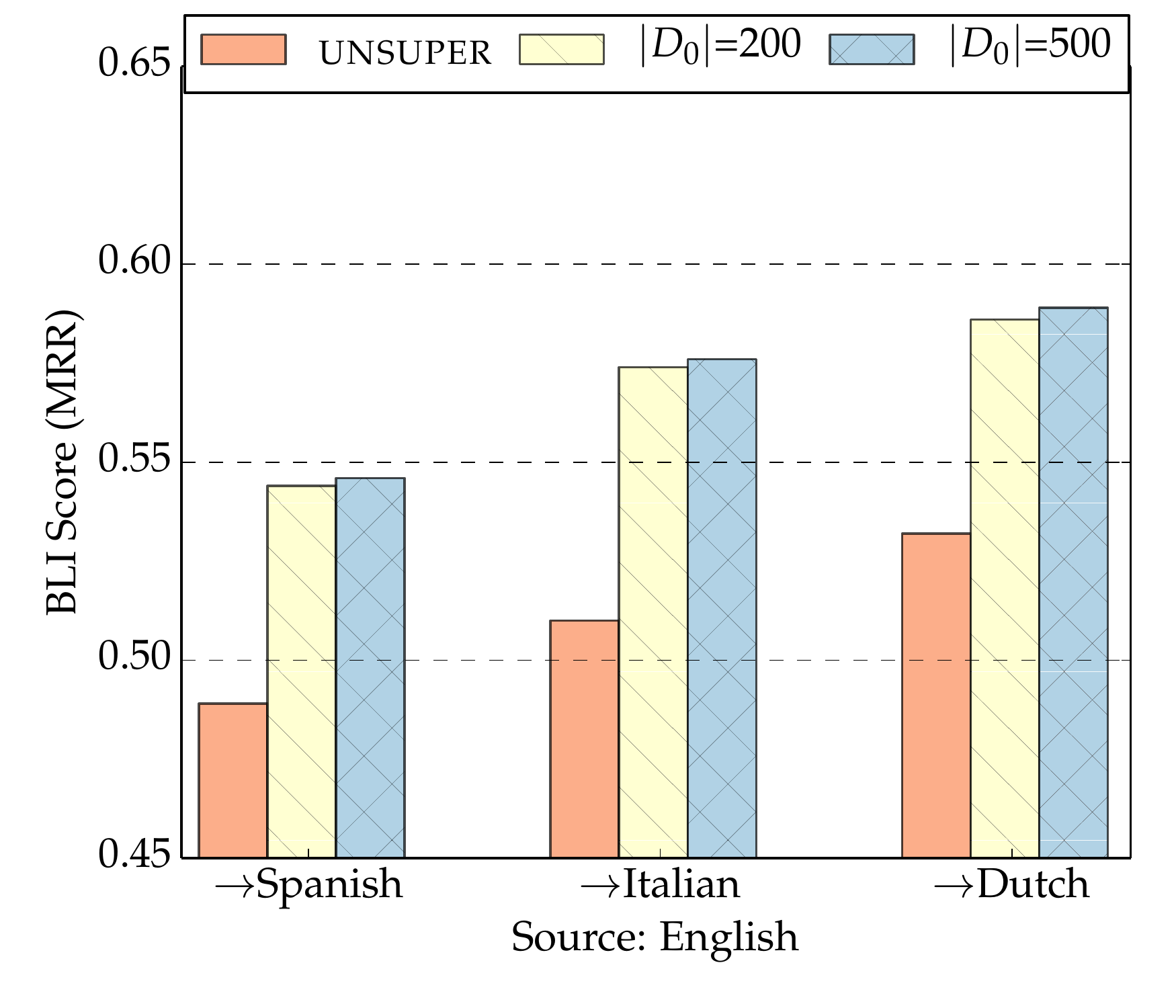}
        \caption{English$\rightarrow L_2$}
        \label{fig:sim-en}
    \end{subfigure}
    \begin{subfigure}[t]{0.328\textwidth}
        \centering
        \includegraphics[width=0.98\linewidth]{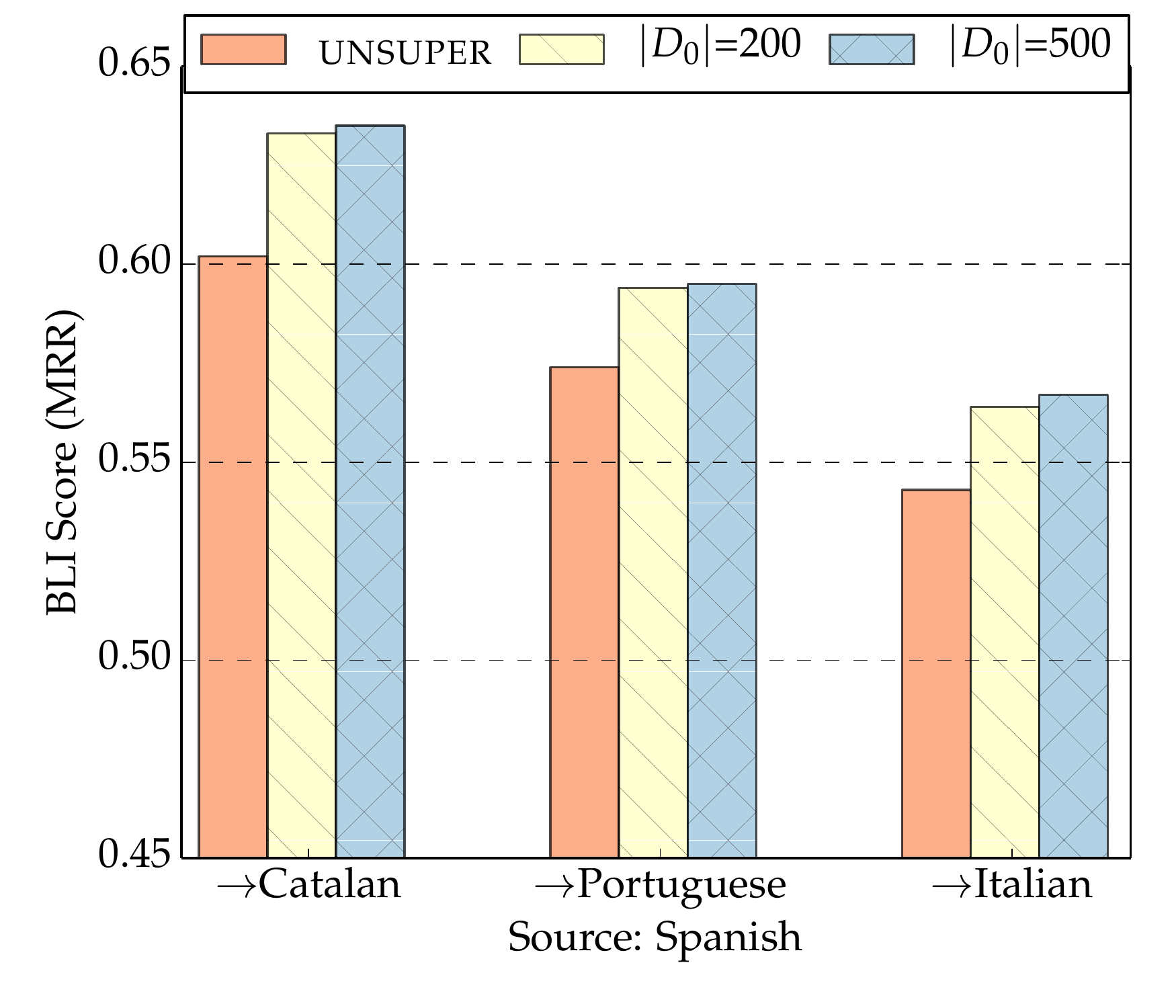}
        \caption{Spanish$\rightarrow L_2$}
        \label{fig:sim-es}
    \end{subfigure}
        \begin{subfigure}[t]{0.323\textwidth}
        \centering
        \includegraphics[width=0.98\linewidth]{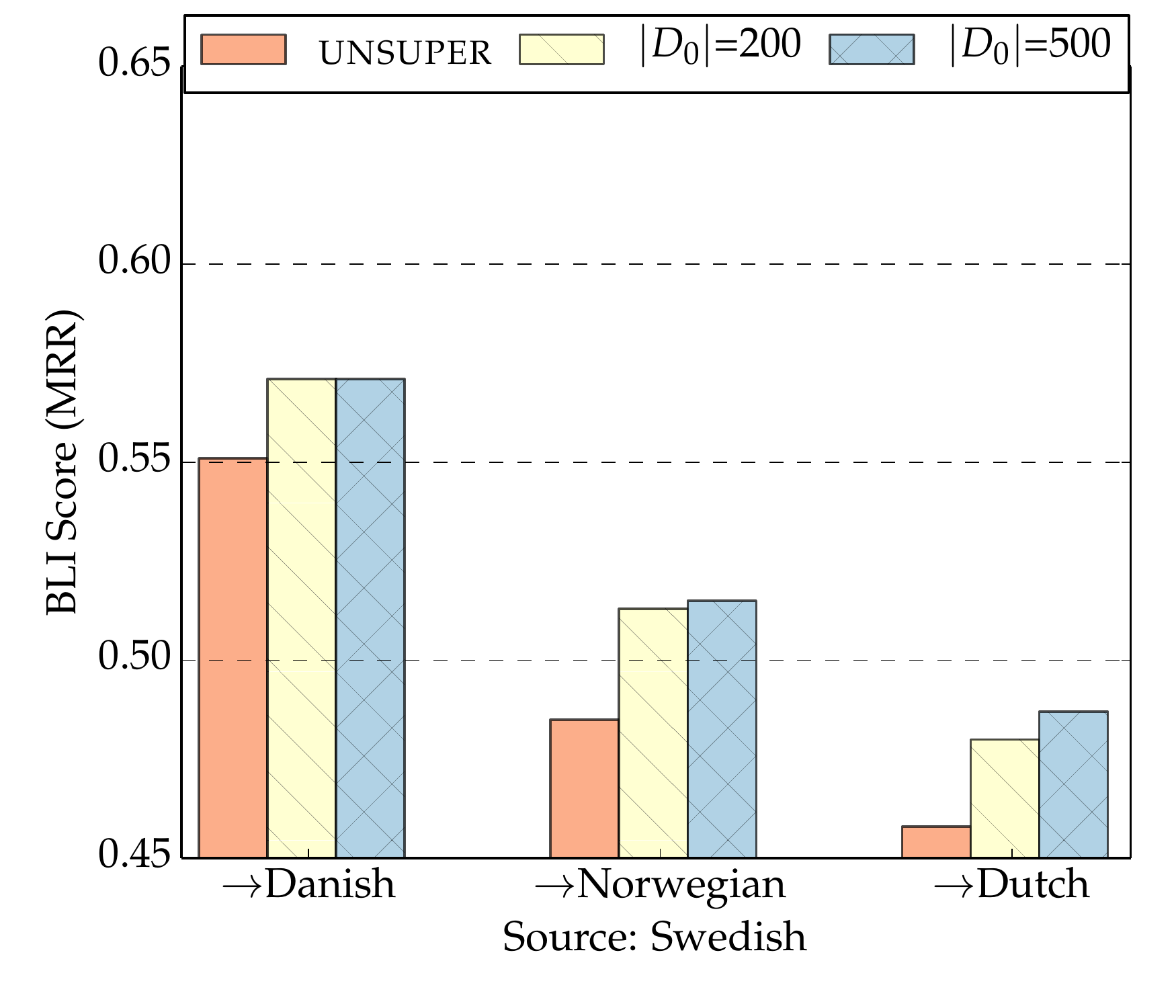}
        \caption{Swedish$\rightarrow L_2$}
        \label{fig:sim-sv}
    \end{subfigure}
    \vspace{-2mm}
    \caption{A comparison of BLI scores on ``easy'' (i.e., similar) language pairs between the fully \textsc{unsupervised} model and a weakly supervised model (seed dictionary size $|D_0|=200$ or $|D_0|=500$) which relies on the self-learning procedure with the symmetry constraint (\textsc{full+sl+sym}).}
    \vspace{-1.5mm}
\label{fig:similar}
\end{figure*}

\begin{figure*}[!t]
    \centering
    \begin{subfigure}[t]{0.328\linewidth}
        \centering
        \includegraphics[width=0.98\linewidth]{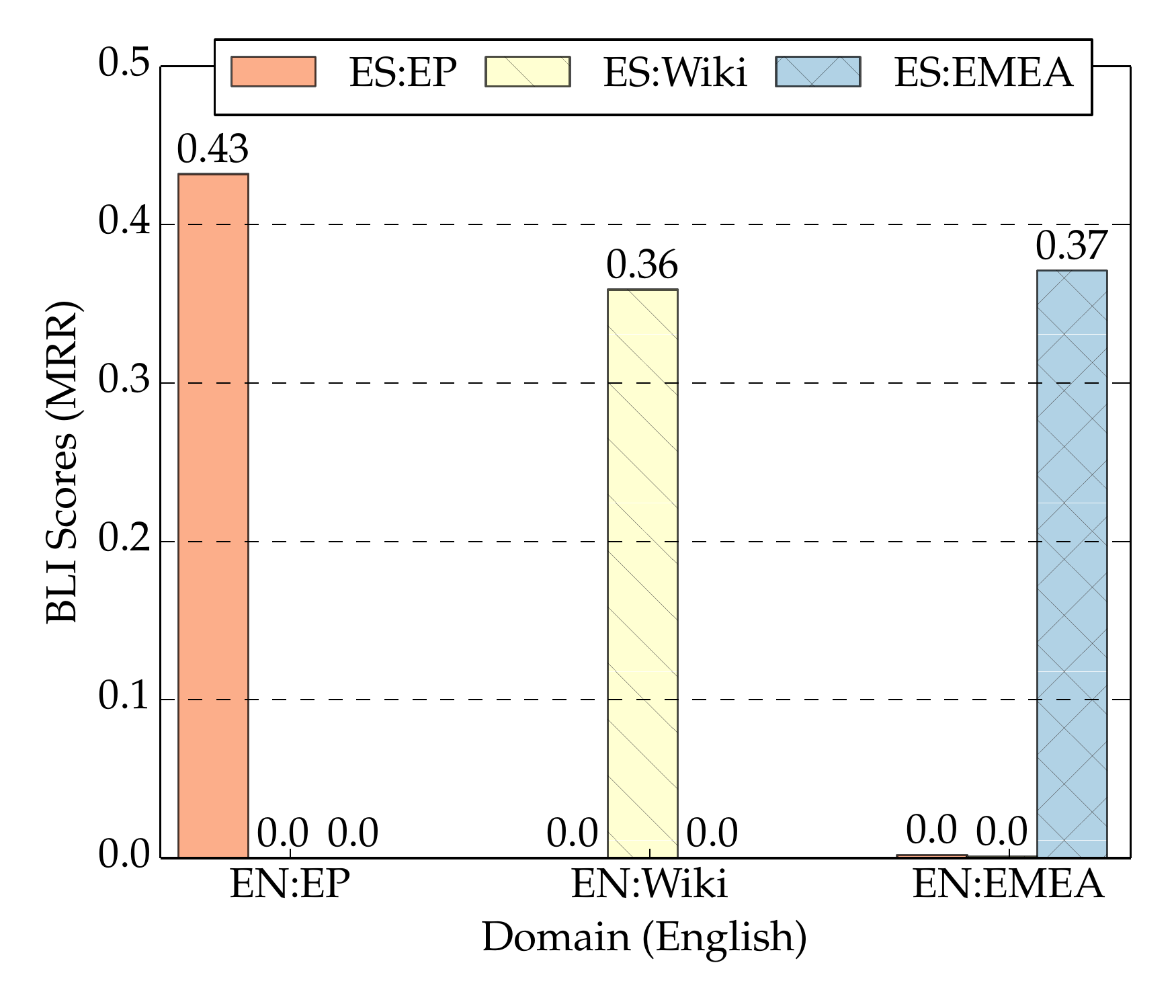}
        \caption{English$\rightarrow$ Spanish}
        \label{fig:domain-en}
    \end{subfigure}
    \begin{subfigure}[t]{0.328\textwidth}
        \centering
        \includegraphics[width=0.98\linewidth]{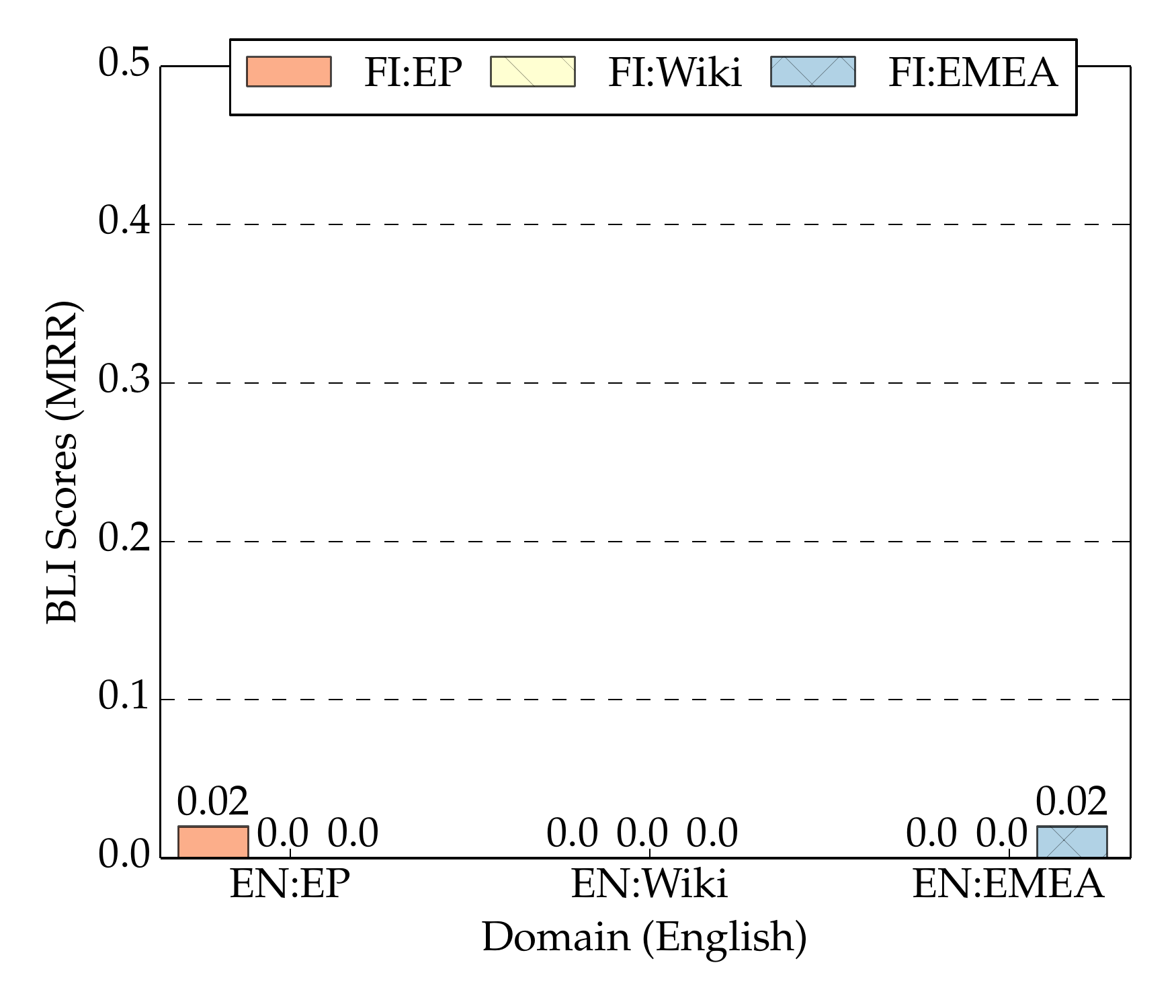}
        \caption{English $\rightarrow$ Finnish}
        \label{fig:domain-fi}
    \end{subfigure}
        \begin{subfigure}[t]{0.323\textwidth}
        \centering
        \includegraphics[width=0.98\linewidth]{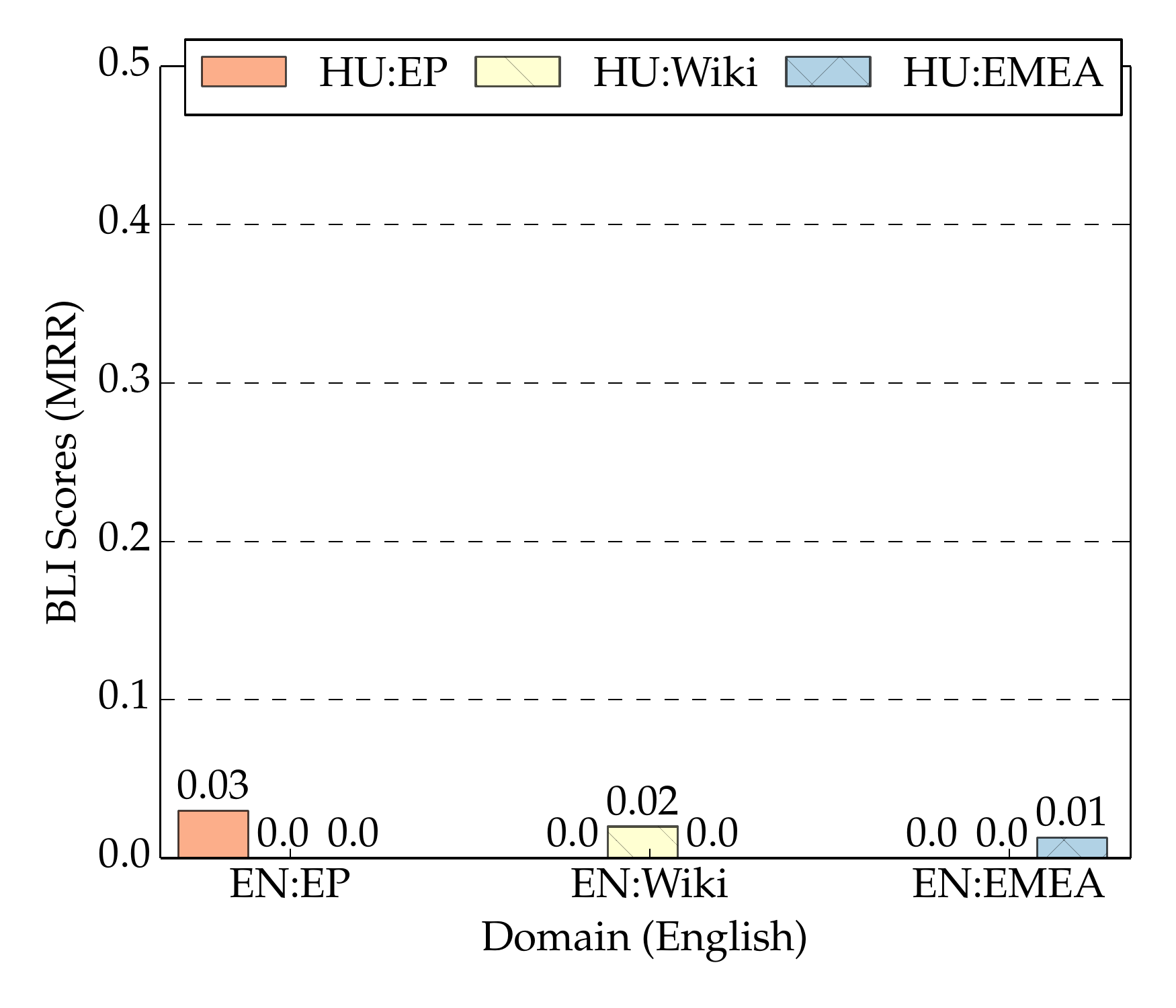}
        \caption{English $\rightarrow$ Hungarian}
        \label{fig:domain-hu}
    \end{subfigure}
    \vspace{-2mm}
    \caption{BLI scores with the (most robust) fully \textsc{unsupervised} model for different language pairs when monolingual word embeddings are trained on dissimilar domains: parliamentary proceedings (EuroParl), Wikipedia (Wiki), and medical corpora (EMEA). Training and test data are the same as in \cite{sogaard2018on}.}
    \vspace{-1.5mm}
\label{fig:domains}
\end{figure*}

\vspace{1.8mm}
\noindent \textbf{How Important are Preprocessing and Postprocessing?} 
The comparisons between \textsc{orthg-super} (and \textsc{orthg+sl+sym}) on the one hand, and \textsc{full-super} (and \textsc{full+sl+sym}) on the other hand clearly indicate that the component C3 plays a substantial role in effective CLWE learning. \textsc{full-super}, which employs all steps S1-S4 (see \S\ref{s:methods}), outperforms \textsc{orthg-super} in 208/210 setups with $|D_0|$=\textit{5k} and in 210/210 setups with $|D_0|$=\textit{1k}. Similarly, \textsc{full+sl+sym} is better than \textsc{orthg+sl+sym} in 210/210 setups (both for $|D_0|$=\textit{1k,5k}). The scores also indicate that dropout with self-learning is useful only when we work with noisy unsupervised seed lexicons: \textsc{full+sl+nod} and \textsc{full+sl+sym} without dropout consistently outperform \textsc{full+sl} across the board.

\vspace{1.8mm}
\noindent \textbf{How Important is (Robust) Self-Learning?}
We note that the best self-learning method is often useful even when $|D_0|=5k$ (i.e., \textsc{full+sl+sym} is better than \textsc{full-super} in 164/210 setups). However, the importance of robust self-learning gets more pronounced as we decrease the size of $D_0$: \textsc{full+sl+sym} is better than \textsc{full-super} in 210/210 setups when $|D_0|=500$ or $|D_0|=1,000$. The gap between the two models, as shown in Figure~\ref{fig:d0}, increases dramatically in favor of \textsc{full+sl+sym} as we decrease $|D_0|$. 

Again, just comparing \textsc{full-super} and \textsc{unsupervised} in Figure~\ref{fig:d0} might give a false impression that fully unsupervised CLWE methods can match their supervised counterparts, but the comparison to \textsc{full+sl+sym} reveals the true extent of performance drop when we abandon even weak supervision. The scores also reveal that the choice of self-learning (C2) does matter: all best performing BLI runs with $|D_0|=1k$ are obtained by two configs with self-learning, and \textsc{full+sl+sym} is the best configuration for 177/210 setups (see Table~\ref{tab:avg-runs}).

\vspace{1.8mm}
\noindent \textbf{Language Pairs.}
As suggested before by \newcite{sogaard2018on} and further verified by \newcite{glavas2019howto} and \newcite{doval2019onthe}, the language pair at hand can have a huge impact on CLWE induction: the adversarial method of \newcite{conneau2018word} often gets stuck in poor local optima and yields degenerate solutions for distant language pairs such as English-Finnish. More recent CLWE methods \cite{artetxe2018robust,Mohiuddin2019revisiting} focus on mitigating this robustness issue. However, they still rely on one critical assumption which leads them to degraded performance for distant language pairs: they assume \textit{approximate isomorphism} \cite{Barone:2016ws,sogaard2018on} between monolingual embedding spaces to learn the initial seed dictionary. In other words, they assume very similar geometric constellations between two monolingual spaces: due to the Zipfian phenomena in language \cite{Zipf:1949book} such near-isomorphism can be satisfied only for \textit{similar languages} and for \textit{similar domains} used for training monolingual vectors. This property is reflected in the results reported in Table~\ref{tab:avg-per-lang}, the number of unsuccessful setups in Table~\ref{tab:avg-runs}, as well as later in Figure~\ref{fig:domains}.

For instance, the largest number of unsuccessful BLI setups with the \textsc{unsupervised} model is reported for Korean, Thai (a tonal language), and Basque (a language isolate): their morphological and genealogical properties are furthest away from other languages in our comparison. A substantial number of unsuccessful setups is also observed with other two language outliers from our set (see Table~\ref{tab:langs} again), Georgian and Indonesian, as well as with morphologically-rich languages such as Estonian or Turkish.


One setting in which fully \textsc{unsupervised} methods did show impressive results in prior work are similar language pairs. However, even in these settings when the comparison to the weakly supervised \textsc{full-super+sym} is completely fair (i.e., the same components C2 and C3 are used for both), \textsc{unsupervised} still falls short of \textsc{full-super+sym}. These results for three source languages are summarized in Figure~\ref{fig:similar}. What is more, one could argue that we do not need unsupervised CLWEs for similar languages in the first place: we can harvest cheap supervision here, e.g., cognates. The main motivation behind unsupervised approaches is to support dissimilar and resource-poor language pairs for which supervision cannot be guaranteed.


\vspace{1.8mm}
\noindent \textbf{Domain Differences.}
Finally, we also verify that \textsc{unsupervised} CLWEs still cannot account for domain differences when training monolingual vectors. We rely on the probing test of \newcite{sogaard2018on}: 300-dim fastText vectors are trained on 1.1M sentences on three corpora: 1) EuroParl.v7 \cite{Koehn:2005} (parliamentary proceedings); 2) Wikipedia \cite{AlRfou:2013conll}, and 3) EMEA \cite{Tiedemann:2009opus} (medical), and BLI evaluation for three language pairs is conducted on standard MUSE BLI test sets \cite{conneau2018word}. The results, summarized in Figure~\ref{fig:domains}, reveal that \textsc{unsupervised} methods are able to yield a good solution only when there is no domain mismatch and for the pair with two most similar languages (English-Spanish), again questioning their robustness and portability to truly low-resource and more challenging setups. Weakly supervised methods ($|D_0|=500$ or $D_0$ seeded with identical strings), in contrast, yield good solutions for all setups.

\section{Further Discussion and Conclusion}
The superiority of weakly supervised methods (e.g., \textsc{full+sl+sym}) over unsupervised methods is especially pronounced for distant and typologically heterogeneous language pairs. However, our study also indicates that even carefully engineered projection-based methods with some seed supervision yield lower absolute performance for such pairs. While we have witnessed the proliferation of fully unsupervised CLWE models recently, some fundamental questions still remain. For instance, the underlying assumption of all projection-based methods (both supervised and unsupervised) is the topological similarity between monolingual spaces, which is why standard simple linear projections result in lower absolute BLI scores for distant pairs (see Table~\ref{tab:avg-runs} and results in the supplemental material). Unsupervised approaches even exploit the assumption \textit{twice} as their seed extraction is fully based on the topological similarity. 


Future work should move beyond the restrictive assumption by exploring new methods that can, e.g., 1) increase the isomorphism between monolingual spaces \cite{Zhang:2019acl} by distinguishing between language-specific and language-pair-invariant subspaces; 2) learn effective non-linear or multiple local projections between monolingual spaces similar to the preliminary work of \newcite{nakashole2018norma}; 3) similar to \newcite{vulic2016on} and \newcite{Lubin:2019naacl} ``denoisify'' seed lexicons during the self-learning procedure. For instance, keeping only mutual/symmetric nearest neighbour as in \textsc{full+sl+sym} can be seen as a form of rudimentary denoisifying: it is indicative to see that the best overall performance in this work is reported with that model configuration. 


Further, the most important contributions of unsupervised CLWE models are, in fact, the improved and more robust self-learning procedures (component C2) and technical enhancements (component C3). In this work we have demonstrated that these components can be equally applied to weakly supervised approaches: starting from a set of only several hundred pairs, they can guarantee consistently improved performance across the board. As there is still no clear-cut use case scenario for unsupervised CLWEs,\footnote{E.g., unsupervised CLWEs are fully substitutable with the superior weakly supervised CLWEs in unsupervised NMT architectures \cite{artetxe2018unsupervised,Lample:2018iclr,lample2018phrase}, or in domain adaptation systems \cite{Ziser2018deep} and fully unsupervised cross-lingual IR \cite{litschko2018unsupervised}.} instead of ``going fully unsupervised'', one pragmatic approach to widen the scope of CLWE learning and its application might invest more effort into extracting at least some seed supervision for a variety of language pairs \cite{artetxe2017learning}. This finding aligns well with the ongoing initiatives of the PanLex project \cite{Kamholz2014panlex} and the ASJP database \cite{Wichmann:2018asjp}, which aim to collate at least some translation pairs in most of the world’s languages.\footnote{\url{https://asjp.clld.org/}}

Finally, this paper demonstrates that, in order to enable fair comparisons, future work on CLWEs should focus on evaluating the CLWE methods' constituent components (e.g, components C1-C3 from this work) instead of full-blown composite systems directly. One goal of the paper is to acknowledge that the work on fully unsupervised CLWE methods has indeed advanced state-of-the-art in cross-lingual word representation learning by offering new solutions also to weakly supervised CLWE methods. However, the robustness problems are still prominent with fully unsupervised CLWEs, and future work should invest more time and effort into developing more robust and more effective methods, e.g., by reaching beyond projection-based methods towards joint approaches \cite{Ruder2018survey,Ormazabal:2019acl}.

\section*{Acknowledgments}
This work is supported by the ERC Consolidator Grant LEXICAL: Lexical Acquisition Across Languages (no 648909). The work of Goran Glavaš is supported by the Baden-Württemberg Stiftung (AGREE grant of the Eliteprogramm). Roi Reichart is partially funded by ISF personal grants No. 1625/18. We thank the three anonymous reviewers for their encouraging comments and suggestions.

\bibliography{refs}
\bibliographystyle{acl_natbib}

\appendix
\clearpage
\section{Supplemental Material}
\label{sec:supplemental}
We report main BLI results for all $15\times 14=210$ language pairs in the supplemental material, grouped by the source language, and for two dictionary sizes: $|D_0|=1,000$ and $|D_0|=500$. The results are provided in Table~\ref{tab:bg-sup}--Table~\ref{tab:tr-sup}. As stressed in the table captions, for the descriptions of benchmarked CLWE configurations, we refer the reader to the main paper: in particular to \S\ref{s:methods} and Table~\ref{tab:config}.

(The actual tables start on the next page.)

\begin{table*}[t]
\def\arraystretch{0.99}
\centering
{\footnotesize
\begin{adjustbox}{max width=\linewidth}
\begin{tabular}{l cccccccccccccc}
\toprule
{} & \multicolumn{14}{c}{\textbf{Bulgarian}: \textsc{bg}-} \\
\cmidrule(lr){2-15}
{} & {-\textsc{ca}} & {-\textsc{eo}} & {-\textsc{et}} & {-\textsc{eu}} & {-\textsc{fi}} & {-\textsc{he}} & {-\textsc{hu}} & {-\textsc{id}} & {-\textsc{ka}} & {-\textsc{ko}} & {-\textsc{lt}} & {-\textsc{no}} & {-\textsc{th}} & {-\textsc{tr}} \\
\cmidrule(lr){2-2} \cmidrule(lr){3-3} \cmidrule(lr){4-4} \cmidrule(lr){5-5} \cmidrule(lr){6-6} \cmidrule(lr){7-7} \cmidrule(lr){8-8} \cmidrule(lr){9-9} \cmidrule(lr){10-10} \cmidrule(lr){11-11} \cmidrule(lr){12-12} \cmidrule(lr){13-13} \cmidrule(lr){14-14} \cmidrule(lr){15-15}
\textsc{unsupervised} & {.353} & {.147} & {.262} & {.005} & {.096} & {.262} & {.386} & {.270} & {.259} & {.004} & {.247} & {.368} & {.001} & {.254} \\
\hdashline
\rowcolor{Gray}
\textsc{5k:orthg-super} & {.350} & {.182} & {.315} & {.173} & {.292} & {.251} & {.379} & {.242} & {.261} & {.125} & {.358} & {.300} & {.089} & {.297} \\
\textsc{5k:orthg+sl+sym} & {.380} & {.215} & {.327} & {.179} & {.319} & {.277} & {.402} & {.271} & {.286} & {.142} & {.372} & {.341} & {.105} & {.321} \\
\rowcolor{Gray}
\textsc{5k:full-super} & {.432} & {.327} & {.407} & {.250} & {.357} & {.361} & {.460} & {.283} & {.364} & {.205} & {.445} & {.398} & {.169} & {.349} \\
\textsc{5k:full+sl} & {.383} & {.273} & {.297} & {.198} & {.203} & {.279} & {.394} & {.277} & {.268} & {.136} & {.301} & {.388} & {.102} & {.291} \\
\rowcolor{Gray}
\textsc{5k:full+sl+nod} & {.427} & {.357} & {.350} & {.245} & {.272} & {.317} & {.439} & {.305} & {.292} & {.179} & {.339} & {.422} & {.154} & {.326} \\
\textsc{5k:full+sl+sym} & {.456} & {.370} & {.405} & {.296} & {.374} & {.368} & {.475} & {.325} & {.367} & {.215} & {.407} & {.446} & {.179} & {.374} \\
\hdashline
\rowcolor{Gray}
\textsc{1k:orthg-super} & {.166} & {.060} & {.141} & {.046} & {.108} & {.076} & {.178} & {.111} & {.098} & {.038} & {.183} & {.122} & {.023} & {.111} \\
\textsc{1k:orthg+sl+sym} & {.310} & {.135} & {.270} & {.087} & {.230} & {.180} & {.346} & {.191} & {.200} & {.071} & {.312} & {.258} & {.030} & {.226} \\
\rowcolor{Gray}
\textsc{1k:full-super} & {.229} & {.147} & {.211} & {.070} & {.129} & {.112} & {.254} & {.116} & {.157} & {.054} & {.230} & {.163} & {.044} & {.133} \\
\textsc{1k:full+sl} & {.381} & {.275} & {.291} & {.198} & {.211} & {.280} & {.392} & {.270} & {.275} & {.127} & {.303} & {.365} & {.101} & {.285} \\
\rowcolor{Gray}
\textsc{1k:full+sl+nod} & {.426} & {.335} & {.347} & {.241} & {.272} & {.315} & {.437} & {.308} & {.306} & {.168} & {.331} & {.413} & {.143} & {.324} \\
\textsc{1k:full+sl+sym} & {.444} & {.357} & {.388} & {.279} & {.361} & {.345} & {.467} & {.314} & {.333} & {.186} & {.369} & {.441} & {.128} & {.357} \\
\bottomrule
\end{tabular}
\end{adjustbox}
}
\vspace{-1.5mm}
\caption{All BLI scores (MRR) with Bulgarian (\textsc{bg}) as the source language. $5k$ and $1k$ denote the seed dictionary $D_0$ size for (weakly) supervised methods. See Table~\ref{tab:config} for a brief description of each model configuration.}
\label{tab:bg-sup}
\end{table*}

\begin{table*}[t]
\def\arraystretch{0.99}
\centering
{\footnotesize
\begin{adjustbox}{max width=\linewidth}
\begin{tabular}{l cccccccccccccc}
\toprule
{} & \multicolumn{14}{c}{\textbf{Catalan}: \textsc{ca}-} \\
\cmidrule(lr){2-15}
{} & {-\textsc{bg}} & {-\textsc{eo}} & {-\textsc{et}} & {-\textsc{eu}} & {-\textsc{fi}} & {-\textsc{he}} & {-\textsc{hu}} & {-\textsc{id}} & {-\textsc{ka}} & {-\textsc{ko}} & {-\textsc{lt}} & {-\textsc{no}} & {-\textsc{th}} & {-\textsc{tr}} \\
\cmidrule(lr){2-2} \cmidrule(lr){3-3} \cmidrule(lr){4-4} \cmidrule(lr){5-5} \cmidrule(lr){6-6} \cmidrule(lr){7-7} \cmidrule(lr){8-8} \cmidrule(lr){9-9} \cmidrule(lr){10-10} \cmidrule(lr){11-11} \cmidrule(lr){12-12} \cmidrule(lr){13-13} \cmidrule(lr){14-14} \cmidrule(lr){15-15}
\textsc{unsupervised} & {.330} & {.310} & {.198} & {.258} & {.228} & {.239} & {.334} & {.302} & {.177} & {.002} & {.163} & {.347} & {.000} & {.260} \\
\hdashline
\rowcolor{Gray}
\textsc{5k:orthg-super} & {.315} & {.199} & {.244} & {.219} & {.252} & {.227} & {.360} & {.249} & {.199} & {.130} & {.258} & {.294} & {.093} & {.283} \\
\textsc{5k:orthg+sl+sym} & {.331} & {.239} & {.269} & {.244} & {.288} & {.258} & {.390} & {.280} & {.213} & {.149} & {.282} & {.340} & {.108} & {.301} \\
\rowcolor{Gray}
\textsc{5k:full-super} & {.396} & {.395} & {.356} & {.338} & {.329} & {.336} & {.431} & {.286} & {.309} & {.217} & {.366} & {.396} & {.196} & {.337} \\
\textsc{5k:full+sl} & {.331} & {.357} & {.223} & {.289} & {.214} & {.257} & {.368} & {.299} & {.174} & {.128} & {.215} & {.366} & {.171} & {.275} \\
\rowcolor{Gray}
\textsc{5k:full+sl+nod} & {.372} & {.458} & {.276} & {.347} & {.265} & {.290} & {.411} & {.317} & {.224} & {.180} & {.264} & {.416} & {.223} & {.308} \\
\textsc{5k:full+sl+sym} & {.414} & {.456} & {.352} & {.391} & {.356} & {.357} & {.449} & {.322} & {.302} & {.245} & {.343} & {.433} & {.218} & {.348} \\
\hdashline
\rowcolor{Gray}
\textsc{1k:orthg-super} & {.160} & {.075} & {.089} & {.063} & {.098} & {.064} & {.151} & {.117} & {.052} & {.029} & {.098} & {.101} & {.018} & {.116} \\
\textsc{1k:orthg+sl+sym} & {.274} & {.151} & {.174} & {.126} & {.168} & {.143} & {.298} & {.207} & {.096} & {.047} & {.198} & {.235} & {.025} & {.193} \\
\rowcolor{Gray}
\textsc{1k:full-super} & {.212} & {.167} & {.165} & {.116} & {.110} & {.103} & {.210} & {.126} & {.101} & {.046} & {.144} & {.138} & {.035} & {.133} \\
\textsc{1k:full+sl} & {.329} & {.361} & {.223} & {.282} & {.218} & {.252} & {.366} & {.299} & {.166} & {.129} & {.225} & {.358} & {.162} & {.272} \\
\rowcolor{Gray}
\textsc{1k:full+sl+nod} & {.372} & {.457} & {.274} & {.342} & {.263} & {.289} & {.411} & {.316} & {.213} & {.170} & {.256} & {.416} & {.217} & {.308} \\
\textsc{1k:full+sl+sym} & {.395} & {.446} & {.300} & {.370} & {.319} & {.335} & {.435} & {.320} & {.253} & {.202} & {.295} & {.424} & {.142} & {.334} \\
\bottomrule
\end{tabular}
\end{adjustbox}
}
\vspace{-1.5mm}
\caption{All BLI scores (MRR) with Catalan (\textsc{ca}) as the source language. $5k$ and $1k$ denote the seed dictionary $D_0$ size for (weakly) supervised methods. See Table~\ref{tab:config} for a brief description of each model configuration.}
\label{tab:ca-sup}
\end{table*}

\begin{table*}[t]
\def\arraystretch{0.99}
\centering
{\footnotesize
\begin{adjustbox}{max width=\linewidth}
\begin{tabular}{l cccccccccccccc}
\toprule
{} & \multicolumn{14}{c}{\textbf{Esperanto}: \textsc{eo}-} \\
\cmidrule(lr){2-15}
{} & {-\textsc{bg}} & {-\textsc{ca}} & {-\textsc{et}} & {-\textsc{eu}} & {-\textsc{fi}} & {-\textsc{he}} & {-\textsc{hu}} & {-\textsc{id}} & {-\textsc{ka}} & {-\textsc{ko}} & {-\textsc{lt}} & {-\textsc{no}} & {-\textsc{th}} & {-\textsc{tr}} \\
\cmidrule(lr){2-2} \cmidrule(lr){3-3} \cmidrule(lr){4-4} \cmidrule(lr){5-5} \cmidrule(lr){6-6} \cmidrule(lr){7-7} \cmidrule(lr){8-8} \cmidrule(lr){9-9} \cmidrule(lr){10-10} \cmidrule(lr){11-11} \cmidrule(lr){12-12} \cmidrule(lr){13-13} \cmidrule(lr){14-14} \cmidrule(lr){15-15}
\textsc{unsupervised} & {.164} & {.430} & {.092} & {.147} & {.012} & {.157} & {.040} & {.047} & {.177} & {.000} & {.198} & {.322} & {.000} & {.001} \\
\hdashline
\rowcolor{Gray}
\textsc{5k:orthg-super} & {.247} & {.367} & {.209} & {.169} & {.246} & {.138} & {.319} & {.171} & {.165} & {.082} & {.221} & {.225} & {.046} & {.210} \\
\textsc{5k:orthg+sl+sym} & {.268} & {.401} & {.227} & {.182} & {.264} & {.160} & {.343} & {.189} & {.179} & {.087} & {.246} & {.254} & {.051} & {.220} \\
\rowcolor{Gray}
\textsc{5k:full-super} & {.367} & {.491} & {.334} & {.294} & {.329} & {.258} & {.400} & {.267} & {.281} & {.171} & {.343} & {.337} & {.107} & {.285} \\
\textsc{5k:full+sl} & {.281} & {.486} & {.225} & {.232} & {.237} & {.187} & {.360} & {.270} & {.171} & {.065} & {.215} & {.340} & {.068} & {.220} \\
\rowcolor{Gray}
\textsc{5k:full+sl+nod} & {.367} & {.524} & {.284} & {.273} & {.322} & {.223} & {.409} & {.321} & {.210} & {.161} & {.292} & {.385} & {.103} & {.262} \\
\textsc{5k:full+sl+sym} & {.410} & {.533} & {.342} & {.354} & {.363} & {.288} & {.426} & {.315} & {.296} & {.184} & {.384} & {.390} & {.117} & {.299} \\
\hdashline
\rowcolor{Gray}
\textsc{1k:orthg-super} & {.103} & {.161} & {.064} & {.043} & {.065} & {.022} & {.107} & {.072} & {.038} & {.015} & {.075} & {.061} & {.007} & {.074} \\
\textsc{1k:orthg+sl+sym} & {.163} & {.266} & {.112} & {.074} & {.105} & {.050} & {.186} & {.104} & {.064} & {.021} & {.127} & {.108} & {.008} & {.103} \\
\rowcolor{Gray}
\textsc{1k:full-super} & {.152} & {.221} & {.136} & {.083} & {.080} & {.044} & {.145} & {.099} & {.078} & {.024} & {.120} & {.083} & {.017} & {.087} \\
\textsc{1k:full+sl} & {.289} & {.491} & {.216} & {.250} & {.229} & {.158} & {.357} & {.262} & {.167} & {.065} & {.215} & {.344} & {.066} & {.223} \\
\rowcolor{Gray}
\textsc{1k:full+sl+nod} & {.371} & {.511} & {.272} & {.264} & {.304} & {.194} & {.407} & {.313} & {.205} & {.142} & {.274} & {.383} & {.080} & {.257} \\
\textsc{1k:full+sl+sym} & {.385} & {.521} & {.314} & {.315} & {.328} & {.241} & {.411} & {.298} & {.255} & {.111} & {.358} & {.376} & {.056} & {.259} \\
\bottomrule
\end{tabular}
\end{adjustbox}
}
\vspace{-1.5mm}
\caption{All BLI scores (MRR) with Esperanto (\textsc{eo}) as the source language. $5k$ and $1k$ denote the seed dictionary $D_0$ size for (weakly) supervised methods. See Table~\ref{tab:config} for a brief description of each model configuration.}
\label{tab:eo-sup}
\end{table*}

\begin{table*}[t]
\def\arraystretch{0.99}
\centering
{\footnotesize
\begin{adjustbox}{max width=\linewidth}
\begin{tabular}{l cccccccccccccc}
\toprule
{} & \multicolumn{14}{c}{\textbf{Estonian}: \textsc{et}-} \\
\cmidrule(lr){2-15}
{} & {-\textsc{bg}} & {-\textsc{ca}} & {-\textsc{eo}} & {-\textsc{eu}} & {-\textsc{fi}} & {-\textsc{he}} & {-\textsc{hu}} & {-\textsc{id}} & {-\textsc{ka}} & {-\textsc{ko}} & {-\textsc{lt}} & {-\textsc{no}} & {-\textsc{th}} & {-\textsc{tr}} \\
\cmidrule(lr){2-2} \cmidrule(lr){3-3} \cmidrule(lr){4-4} \cmidrule(lr){5-5} \cmidrule(lr){6-6} \cmidrule(lr){7-7} \cmidrule(lr){8-8} \cmidrule(lr){9-9} \cmidrule(lr){10-10} \cmidrule(lr){11-11} \cmidrule(lr){12-12} \cmidrule(lr){13-13} \cmidrule(lr){14-14} \cmidrule(lr){15-15}
\textsc{unsupervised} & {.283} & {.220} & {.105} & {.004} & {.370} & {.208} & {.377} & {.000} & {.002} & {.001} & {.282} & {.316} & {.000} & {.002} \\
\hdashline
\rowcolor{Gray}
\textsc{5k:orthg-super} & {.295} & {.212} & {.142} & {.171} & {.368} & {.192} & {.342} & {.120} & {.185} & {.112} & {.277} & {.210} & {.076} & {.241} \\
\textsc{5k:orthg+sl+sym} & {.301} & {.234} & {.160} & {.172} & {.388} & {.209} & {.365} & {.136} & {.192} & {.121} & {.286} & {.246} & {.089} & {.255} \\
\rowcolor{Gray}
\textsc{5k:full-super} & {.393} & {.333} & {.271} & {.238} & {.430} & {.287} & {.432} & {.212} & {.258} & {.191} & {.360} & {.328} & {.168} & {.307} \\
\textsc{5k:full+sl} & {.309} & {.258} & {.217} & {.128} & {.393} & {.221} & {.382} & {.160} & {.183} & {.114} & {.286} & {.320} & {.090} & {.249} \\
\rowcolor{Gray}
\textsc{5k:full+sl+nod} & {.354} & {.308} & {.268} & {.181} & {.397} & {.268} & {.418} & {.210} & {.208} & {.152} & {.319} & {.351} & {.150} & {.277} \\
\textsc{5k:full+sl+sym} & {.404} & {.357} & {.307} & {.238} & {.443} & {.301} & {.459} & {.223} & {.251} & {.185} & {.358} & {.383} & {.178} & {.331} \\
\hdashline
\rowcolor{Gray}
\textsc{1k:orthg-super} & {.135} & {.080} & {.060} & {.074} & {.171} & {.050} & {.145} & {.034} & {.067} & {.033} & {.131} & {.069} & {.025} & {.073} \\
\textsc{1k:orthg+sl+sym} & {.238} & {.153} & {.096} & {.108} & {.347} & {.124} & {.307} & {.075} & {.121} & {.060} & {.228} & {.158} & {.033} & {.152} \\
\rowcolor{Gray}
\textsc{1k:full-super} & {.200} & {.121} & {.116} & {.099} & {.200} & {.069} & {.188} & {.065} & {.095} & {.052} & {.179} & {.112} & {.041} & {.102} \\
\textsc{1k:full+sl} & {.289} & {.240} & {.221} & {.127} & {.395} & {.211} & {.380} & {.163} & {.183} & {.117} & {.286} & {.315} & {.074} & {.248} \\
\rowcolor{Gray}
\textsc{1k:full+sl+nod} & {.343} & {.294} & {.267} & {.193} & {.396} & {.259} & {.420} & {.209} & {.204} & {.147} & {.316} & {.352} & {.123} & {.290} \\
\textsc{1k:full+sl+sym} & {.381} & {.346} & {.297} & {.208} & {.437} & {.277} & {.449} & {.204} & {.215} & {.148} & {.337} & {.377} & {.108} & {.313} \\
\bottomrule
\end{tabular}
\end{adjustbox}
}
\vspace{-1.5mm}
\caption{All BLI scores (MRR) with Estonian (\textsc{et}) as the source language. $5k$ and $1k$ denote the seed dictionary $D_0$ size for (weakly) supervised methods. See Table~\ref{tab:config} for a brief description of each model configuration.}
\label{tab:et-sup}
\end{table*}

\begin{table*}[t]
\def\arraystretch{0.99}
\centering
{\footnotesize
\begin{adjustbox}{max width=\linewidth}
\begin{tabular}{l cccccccccccccc}
\toprule
{} & \multicolumn{14}{c}{\textbf{Basque}: \textsc{eu}-} \\
\cmidrule(lr){2-15}
{} & {-\textsc{bg}} & {-\textsc{ca}} & {-\textsc{eo}} & {-\textsc{et}} & {-\textsc{fi}} & {-\textsc{he}} & {-\textsc{hu}} & {-\textsc{id}} & {-\textsc{ka}} & {-\textsc{ko}} & {-\textsc{lt}} & {-\textsc{no}} & {-\textsc{th}} & {-\textsc{tr}} \\
\cmidrule(lr){2-2} \cmidrule(lr){3-3} \cmidrule(lr){4-4} \cmidrule(lr){5-5} \cmidrule(lr){6-6} \cmidrule(lr){7-7} \cmidrule(lr){8-8} \cmidrule(lr){9-9} \cmidrule(lr){10-10} \cmidrule(lr){11-11} \cmidrule(lr){12-12} \cmidrule(lr){13-13} \cmidrule(lr){14-14} \cmidrule(lr){15-15}
\textsc{unsupervised} & {.002} & {.254} & {.115} & {.007} & {.008} & {.048} & {.001} & {.001} & {.059} & {.000} & {.001} & {.000} & {.000} & {.000} \\
\hdashline
\rowcolor{Gray}
\textsc{5k:orthg-super} & {.190} & {.279} & {.122} & {.174} & {.155} & {.134} & {.178} & {.134} & {.154} & {.069} & {.159} & {.138} & {.045} & {.176} \\
\textsc{5k:orthg+sl+sym} & {.196} & {.307} & {.135} & {.181} & {.175} & {.148} & {.193} & {.147} & {.165} & {.076} & {.175} & {.155} & {.049} & {.187} \\
\rowcolor{Gray}
\textsc{5k:full-super} & {.292} & {.391} & {.245} & {.250} & {.233} & {.211} & {.259} & {.183} & {.197} & {.109} & {.242} & {.240} & {.095} & {.240} \\
\textsc{5k:full+sl} & {.203} & {.371} & {.203} & {.126} & {.175} & {.110} & {.187} & {.209} & {.084} & {.043} & {.098} & {.201} & {.069} & {.170} \\
\rowcolor{Gray}
\textsc{5k:full+sl+nod} & {.248} & {.423} & {.243} & {.181} & {.219} & {.163} & {.228} & {.244} & {.121} & {.081} & {.148} & {.252} & {.093} & {.208} \\
\textsc{5k:full+sl+sym} & {.310} & {.441} & {.277} & {.248} & {.270} & {.206} & {.283} & {.225} & {.189} & {.106} & {.237} & {.287} & {.094} & {.248} \\
\hdashline
\rowcolor{Gray}
\textsc{1k:orthg-super} & {.090} & {.117} & {.038} & {.061} & {.045} & {.029} & {.054} & {.044} & {.051} & {.019} & {.052} & {.037} & {.012} & {.044} \\
\textsc{1k:orthg+sl+sym} & {.116} & {.192} & {.061} & {.105} & {.072} & {.052} & {.102} & {.062} & {.080} & {.029} & {.086} & {.066} & {.013} & {.077} \\
\rowcolor{Gray}
\textsc{1k:full-super} & {.120} & {.142} & {.077} & {.088} & {.048} & {.037} & {.077} & {.049} & {.059} & {.021} & {.071} & {.053} & {.018} & {.055} \\
\textsc{1k:full+sl} & {.206} & {.348} & {.207} & {.117} & {.172} & {.115} & {.186} & {.209} & {.080} & {.039} & {.101} & {.201} & {.072} & {.159} \\
\rowcolor{Gray}
\textsc{1k:full+sl+nod} & {.244} & {.418} & {.230} & {.179} & {.214} & {.163} & {.225} & {.236} & {.111} & {.073} & {.136} & {.248} & {.077} & {.202} \\
\textsc{1k:full+sl+sym} & {.276} & {.428} & {.253} & {.213} & {.247} & {.166} & {.266} & {.213} & {.147} & {.060} & {.169} & {.261} & {.056} & {.212} \\
\bottomrule
\end{tabular}
\end{adjustbox}
}
\vspace{-1.5mm}
\caption{All BLI scores (MRR) with Basque (\textsc{eu}) as the source language. $5k$ and $1k$ denote the seed dictionary $D_0$ size for (weakly) supervised methods. See Table~\ref{tab:config} for a brief description of each model configuration.}
\label{tab:eu-sup}
\end{table*}

\begin{table*}[t]
\def\arraystretch{0.99}
\centering
{\footnotesize
\begin{adjustbox}{max width=\linewidth}
\begin{tabular}{l cccccccccccccc}
\toprule
{} & \multicolumn{14}{c}{\textbf{Finnish}: \textsc{fi}-} \\
\cmidrule(lr){2-15}
{} & {-\textsc{bg}} & {-\textsc{ca}} & {-\textsc{eo}} & {-\textsc{et}} & {-\textsc{eu}} & {-\textsc{he}} & {-\textsc{hu}} & {-\textsc{id}} & {-\textsc{ka}} & {-\textsc{ko}} & {-\textsc{lt}} & {-\textsc{no}} & {-\textsc{th}} & {-\textsc{tr}} \\
\cmidrule(lr){2-2} \cmidrule(lr){3-3} \cmidrule(lr){4-4} \cmidrule(lr){5-5} \cmidrule(lr){6-6} \cmidrule(lr){7-7} \cmidrule(lr){8-8} \cmidrule(lr){9-9} \cmidrule(lr){10-10} \cmidrule(lr){11-11} \cmidrule(lr){12-12} \cmidrule(lr){13-13} \cmidrule(lr){14-14} \cmidrule(lr){15-15}
\textsc{unsupervised} & {.120} & {.304} & {.013} & {.361} & {.004} & {.243} & {.411} & {.230} & {.003} & {.000} & {.284} & {.395} & {.000} & {.162} \\
\hdashline
\rowcolor{Gray}
\textsc{5k:orthg-super} & {.292} & {.295} & {.132} & {.302} & {.116} & {.216} & {.407} & {.221} & {.166} & {.132} & {.265} & {.332} & {.082} & {.300} \\
\textsc{5k:orthg+sl+sym} & {.311} & {.326} & {.147} & {.336} & {.130} & {.245} & {.428} & {.243} & {.185} & {.150} & {.292} & {.370} & {.089} & {.325} \\
\rowcolor{Gray}
\textsc{5k:full-super} & {.379} & {.377} & {.284} & {.409} & {.220} & {.323} & {.456} & {.263} & {.275} & {.222} & {.390} & {.419} & {.171} & {.346} \\
\textsc{5k:full+sl} & {.269} & {.325} & {.216} & {.360} & {.153} & {.245} & {.408} & {.260} & {.189} & {.136} & {.280} & {.404} & {.093} & {.303} \\
\rowcolor{Gray}
\textsc{5k:full+sl+nod} & {.353} & {.388} & {.299} & {.401} & {.205} & {.311} & {.453} & {.296} & {.251} & {.196} & {.372} & {.449} & {.165} & {.335} \\
\textsc{5k:full+sl+sym} & {.397} & {.404} & {.320} & {.424} & {.271} & {.351} & {.474} & {.298} & {.289} & {.243} & {.405} & {.460} & {.168} & {.365} \\
\hdashline
\rowcolor{Gray}
\textsc{1k:orthg-super} & {.134} & {.114} & {.033} & {.105} & {.030} & {.060} & {.195} & {.091} & {.047} & {.035} & {.113} & {.140} & {.014} & {.128} \\
\textsc{1k:orthg+sl+sym} & {.246} & {.198} & {.049} & {.275} & {.047} & {.129} & {.368} & {.141} & {.086} & {.059} & {.228} & {.285} & {.017} & {.230} \\
\rowcolor{Gray}
\textsc{1k:full-super} & {.174} & {.142} & {.077} & {.167} & {.054} & {.071} & {.226} & {.098} & {.084} & {.052} & {.158} & {.161} & {.028} & {.149} \\
\textsc{1k:full+sl} & {.258} & {.276} & {.222} & {.361} & {.154} & {.253} & {.400} & {.255} & {.199} & {.129} & {.309} & {.402} & {.065} & {.309} \\
\rowcolor{Gray}
\textsc{1k:full+sl+nod} & {.351} & {.330} & {.298} & {.400} & {.193} & {.309} & {.452} & {.299} & {.246} & {.183} & {.372} & {.448} & {.141} & {.334} \\
\textsc{1k:full+sl+sym} & {.381} & {.396} & {.304} & {.416} & {.235} & {.331} & {.463} & {.300} & {.270} & {.211} & {.389} & {.455} & {.107} & {.353} \\
\bottomrule
\end{tabular}
\end{adjustbox}
}
\vspace{-1.5mm}
\caption{All BLI scores (MRR) with Finnish (\textsc{fi}) as the source language. $5k$ and $1k$ denote the seed dictionary $D_0$ size for (weakly) supervised methods. See Table~\ref{tab:config} for a brief description of each model configuration.}
\label{tab:fi-sup}
\end{table*}

\begin{table*}[t]
\def\arraystretch{0.99}
\centering
{\footnotesize
\begin{adjustbox}{max width=\linewidth}
\begin{tabular}{l cccccccccccccc}
\toprule
{} & \multicolumn{14}{c}{\textbf{Hebrew}: \textsc{he}-} \\
\cmidrule(lr){2-15}
{} & {-\textsc{bg}} & {-\textsc{ca}} & {-\textsc{eo}} & {-\textsc{et}} & {-\textsc{eu}} & {-\textsc{fi}} & {-\textsc{hu}} & {-\textsc{id}} & {-\textsc{ka}} & {-\textsc{ko}} & {-\textsc{lt}} & {-\textsc{no}} & {-\textsc{th}} & {-\textsc{tr}} \\
\cmidrule(lr){2-2} \cmidrule(lr){3-3} \cmidrule(lr){4-4} \cmidrule(lr){5-5} \cmidrule(lr){6-6} \cmidrule(lr){7-7} \cmidrule(lr){8-8} \cmidrule(lr){9-9} \cmidrule(lr){10-10} \cmidrule(lr){11-11} \cmidrule(lr){12-12} \cmidrule(lr){13-13} \cmidrule(lr){14-14} \cmidrule(lr){15-15}
\textsc{unsupervised} & {.297} & {.319} & {.172} & {.180} & {.101} & {.214} & {.318} & {.212} & {.123} & {.004} & {.106} & {.292} & {.000} & {.262} \\
\hdashline
\rowcolor{Gray}
\textsc{5k:orthg-super} & {.288} & {.269} & {.135} & {.193} & {.160} & {.250} & {.287} & {.154} & {.161} & {.126} & {.214} & {.205} & {.086} & {.247} \\
\textsc{5k:orthg+sl+sym} & {.307} & {.298} & {.149} & {.206} & {.161} & {.279} & {.315} & {.181} & {.167} & {.135} & {.229} & {.238} & {.107} & {.274} \\
\rowcolor{Gray}
\textsc{5k:full-super} & {.397} & {.376} & {.248} & {.288} & {.225} & {.329} & {.375} & {.239} & {.213} & {.204} & {.309} & {.316} & {.173} & {.328} \\
\textsc{5k:full+sl} & {.304} & {.316} & {.193} & {.187} & {.137} & {.223} & {.314} & {.212} & {.142} & {.142} & {.193} & {.284} & {.124} & {.267} \\
\rowcolor{Gray}
\textsc{5k:full+sl+nod} & {.343} & {.356} & {.241} & {.230} & {.171} & {.269} & {.361} & {.250} & {.179} & {.185} & {.237} & {.323} & {.190} & {.298} \\
\textsc{5k:full+sl+sym} & {.378} & {.384} & {.278} & {.278} & {.211} & {.320} & {.393} & {.266} & {.217} & {.218} & {.301} & {.349} & {.192} & {.337} \\
\hdashline
\rowcolor{Gray}
\textsc{1k:orthg-super} & {.126} & {.101} & {.040} & {.064} & {.050} & {.068} & {.096} & {.057} & {.048} & {.033} & {.078} & {.058} & {.022} & {.086} \\
\textsc{1k:orthg+sl+sym} & {.243} & {.204} & {.079} & {.134} & {.097} & {.144} & {.199} & {.116} & {.098} & {.061} & {.154} & {.130} & {.031} & {.170} \\
\rowcolor{Gray}
\textsc{1k:full-super} & {.180} & {.148} & {.087} & {.106} & {.065} & {.077} & {.135} & {.076} & {.067} & {.054} & {.105} & {.086} & {.042} & {.111} \\
\textsc{1k:full+sl} & {.309} & {.317} & {.195} & {.191} & {.141} & {.226} & {.319} & {.209} & {.138} & {.140} & {.189} & {.288} & {.105} & {.271} \\
\rowcolor{Gray}
\textsc{1k:full+sl+nod} & {.345} & {.353} & {.235} & {.226} & {.162} & {.268} & {.360} & {.250} & {.176} & {.183} & {.232} & {.317} & {.159} & {.304} \\
\textsc{1k:full+sl+sym} & {.360} & {.371} & {.252} & {.250} & {.182} & {.293} & {.383} & {.251} & {.188} & {.187} & {.254} & {.343} & {.114} & {.321} \\
\bottomrule
\end{tabular}
\end{adjustbox}
}
\vspace{-1.5mm}
\caption{All BLI scores (MRR) with Hebrew (\textsc{he}) as the source language. $5k$ and $1k$ denote the seed dictionary $D_0$ size for (weakly) supervised methods. See Table~\ref{tab:config} for a brief description of each model configuration.}
\label{tab:he-sup}
\end{table*}

\begin{table*}[t]
\def\arraystretch{0.99}
\centering
{\footnotesize
\begin{adjustbox}{max width=\linewidth}
\begin{tabular}{l cccccccccccccc}
\toprule
{} & \multicolumn{14}{c}{\textbf{Hungarian}: \textsc{hu}-} \\
\cmidrule(lr){2-15}
{} & {-\textsc{bg}} & {-\textsc{ca}} & {-\textsc{eo}} & {-\textsc{et}} & {-\textsc{eu}} & {-\textsc{fi}} & {-\textsc{he}} & {-\textsc{id}} & {-\textsc{ka}} & {-\textsc{ko}} & {-\textsc{lt}} & {-\textsc{no}} & {-\textsc{th}} & {-\textsc{tr}} \\
\cmidrule(lr){2-2} \cmidrule(lr){3-3} \cmidrule(lr){4-4} \cmidrule(lr){5-5} \cmidrule(lr){6-6} \cmidrule(lr){7-7} \cmidrule(lr){8-8} \cmidrule(lr){9-9} \cmidrule(lr){10-10} \cmidrule(lr){11-11} \cmidrule(lr){12-12} \cmidrule(lr){13-13} \cmidrule(lr){14-14} \cmidrule(lr){15-15}
\textsc{unsupervised} & {.328} & {.362} & {.012} & {.298} & {.001} & {.342} & {.275} & {.318} & {.027} & {.000} & {.232} & {.359} & {.000} & {.330} \\
\hdashline
\rowcolor{Gray}
\textsc{5k:orthg-super} & {.336} & {.360} & {.198} & {.309} & {.191} & {.339} & {.227} & {.254} & {.218} & {.158} & {.294} & {.310} & {.096} & {.331} \\
\textsc{5k:orthg+sl+sym} & {.357} & {.391} & {.222} & {.341} & {.204} & {.372} & {.262} & {.277} & {.237} & {.184} & {.326} & {.345} & {.105} & {.348} \\
\rowcolor{Gray}
\textsc{5k:full-super} & {.431} & {.443} & {.344} & {.423} & {.282} & {.397} & {.349} & {.338} & {.326} & {.259} & {.411} & {.406} & {.173} & {.372} \\
\textsc{5k:full+sl} & {.348} & {.402} & {.302} & {.325} & {.205} & {.343} & {.275} & {.318} & {.218} & {.187} & {.246} & {.361} & {.087} & {.338} \\
\rowcolor{Gray}
\textsc{5k:full+sl+nod} & {.397} & {.445} & {.378} & {.370} & {.252} & {.372} & {.319} & {.364} & {.265} & {.228} & {.319} & {.402} & {.145} & {.361} \\
\textsc{5k:full+sl+sym} & {.438} & {.477} & {.392} & {.433} & {.305} & {.407} & {.376} & {.374} & {.332} & {.285} & {.419} & {.441} & {.176} & {.380} \\
\hdashline
\rowcolor{Gray}
\textsc{1k:orthg-super} & {.164} & {.170} & {.063} & {.125} & {.056} & {.149} & {.068} & {.113} & {.063} & {.039} & {.128} & {.102} & {.008} & {.161} \\
\textsc{1k:orthg+sl+sym} & {.297} & {.298} & {.107} & {.255} & {.091} & {.293} & {.152} & {.200} & {.129} & {.076} & {.237} & {.243} & {.016} & {.287} \\
\rowcolor{Gray}
\textsc{1k:full-super} & {.241} & {.221} & {.125} & {.196} & {.094} & {.168} & {.098} & {.147} & {.112} & {.063} & {.183} & {.149} & {.026} & {.184} \\
\textsc{1k:full+sl} & {.337} & {.403} & {.292} & {.312} & {.217} & {.341} & {.256} & {.317} & {.219} & {.186} & {.248} & {.360} & {.086} & {.334} \\
\rowcolor{Gray}
\textsc{1k:full+sl+nod} & {.395} & {.447} & {.374} & {.364} & {.260} & {.375} & {.312} & {.359} & {.264} & {.231} & {.317} & {.398} & {.132} & {.355} \\
\textsc{1k:full+sl+sym} & {.427} & {.467} & {.369} & {.413} & {.274} & {.400} & {.356} & {.377} & {.306} & {.268} & {.381} & {.423} & {.113} & {.374} \\
\bottomrule
\end{tabular}
\end{adjustbox}
}
\vspace{-1.5mm}
\caption{All BLI scores (MRR) with Hungarian (\textsc{hu}) as the source language. $5k$ and $1k$ denote the seed dictionary $D_0$ size for (weakly) supervised methods. See Table~\ref{tab:config} for a brief description of each model configuration.}
\label{tab:hu-sup}
\end{table*}

\begin{table*}[t]
\def\arraystretch{0.99}
\centering
{\footnotesize
\begin{adjustbox}{max width=\linewidth}
\begin{tabular}{l cccccccccccccc}
\toprule
{} & \multicolumn{14}{c}{\textbf{Indonesian}: \textsc{id}-} \\
\cmidrule(lr){2-15}
{} & {-\textsc{bg}} & {-\textsc{ca}} & {-\textsc{eo}} & {-\textsc{et}} & {-\textsc{eu}} & {-\textsc{fi}} & {-\textsc{he}} & {-\textsc{hu}} & {-\textsc{ka}} & {-\textsc{ko}} & {-\textsc{lt}} & {-\textsc{no}} & {-\textsc{th}} & {-\textsc{tr}} \\
\cmidrule(lr){2-2} \cmidrule(lr){3-3} \cmidrule(lr){4-4} \cmidrule(lr){5-5} \cmidrule(lr){6-6} \cmidrule(lr){7-7} \cmidrule(lr){8-8} \cmidrule(lr){9-9} \cmidrule(lr){10-10} \cmidrule(lr){11-11} \cmidrule(lr){12-12} \cmidrule(lr){13-13} \cmidrule(lr){14-14} \cmidrule(lr){15-15}
\textsc{unsupervised} & {.132} & {.271} & {.005} & {.000} & {.001} & {.174} & {.190} & {.256} & {.001} & {.001} & {.000} & {.257} & {.001} & {.252} \\
\hdashline
\rowcolor{Gray}
\textsc{5k:orthg-super} & {.223} & {.256} & {.153} & {.217} & {.111} & {.182} & {.222} & {.273} & {.162} & {.126} & {.201} & {.233} & {.147} & {.276} \\
\textsc{5k:orthg+sl+sym} & {.237} & {.278} & {.167} & {.228} & {.123} & {.219} & {.239} & {.294} & {.170} & {.140} & {.207} & {.265} & {.163} & {.294} \\
\rowcolor{Gray}
\textsc{5k:full-super} & {.281} & {.300} & {.247} & {.281} & {.173} & {.233} & {.290} & {.349} & {.222} & {.193} & {.260} & {.294} & {.218} & {.316} \\
\textsc{5k:full+sl} & {.156} & {.273} & {.170} & {.126} & {.132} & {.171} & {.201} & {.256} & {.099} & {.126} & {.073} & {.258} & {.215} & {.265} \\
\rowcolor{Gray}
\textsc{5k:full+sl+nod} & {.210} & {.300} & {.218} & {.174} & {.166} & {.226} & {.238} & {.300} & {.133} & {.172} & {.110} & {.288} & {.253} & {.292} \\
\textsc{5k:full+sl+sym} & {.287} & {.323} & {.274} & {.266} & {.220} & {.269} & {.295} & {.345} & {.200} & {.197} & {.242} & {.320} & {.241} & {.326} \\
\hdashline
\rowcolor{Gray}
\textsc{1k:orthg-super} & {.092} & {.098} & {.050} & {.075} & {.025} & {.053} & {.068} & {.109} & {.039} & {.043} & {.069} & {.074} & {.037} & {.129} \\
\textsc{1k:orthg+sl+sym} & {.150} & {.168} & {.078} & {.142} & {.037} & {.082} & {.140} & {.197} & {.071} & {.078} & {.108} & {.142} & {.052} & {.225} \\
\rowcolor{Gray}
\textsc{1k:full-super} & {.121} & {.114} & {.092} & {.115} & {.038} & {.053} & {.093} & {.129} & {.063} & {.062} & {.086} & {.081} & {.052} & {.152} \\
\textsc{1k:full+sl} & {.149} & {.267} & {.170} & {.132} & {.134} & {.171} & {.198} & {.257} & {.101} & {.123} & {.073} & {.257} & {.217} & {.266} \\
\rowcolor{Gray}
\textsc{1k:full+sl+nod} & {.208} & {.301} & {.213} & {.167} & {.165} & {.218} & {.240} & {.299} & {.126} & {.165} & {.101} & {.287} & {.246} & {.290} \\
\textsc{1k:full+sl+sym} & {.258} & {.316} & {.254} & {.213} & {.187} & {.250} & {.264} & {.337} & {.140} & {.175} & {.152} & {.309} & {.226} & {.319} \\
\bottomrule
\end{tabular}
\end{adjustbox}
}
\vspace{-1.5mm}
\caption{All BLI scores (MRR) with Indonesian (\textsc{id}) as the source language. $5k$ and $1k$ denote the seed dictionary $D_0$ size for (weakly) supervised methods. See Table~\ref{tab:config} for a brief description of each model configuration.}
\label{tab:id-sup}
\end{table*}

\begin{table*}[t]
\def\arraystretch{0.99}
\centering
{\footnotesize
\begin{adjustbox}{max width=\linewidth}
\begin{tabular}{l cccccccccccccc}
\toprule
{} & \multicolumn{14}{c}{\textbf{Georgian}: \textsc{ka}-} \\
\cmidrule(lr){2-15}
{} & {-\textsc{bg}} & {-\textsc{ca}} & {-\textsc{eo}} & {-\textsc{et}} & {-\textsc{eu}} & {-\textsc{fi}} & {-\textsc{he}} & {-\textsc{hu}} & {-\textsc{id}} & {-\textsc{ko}} & {-\textsc{lt}} & {-\textsc{no}} & {-\textsc{th}} & {-\textsc{tr}} \\
\cmidrule(lr){2-2} \cmidrule(lr){3-3} \cmidrule(lr){4-4} \cmidrule(lr){5-5} \cmidrule(lr){6-6} \cmidrule(lr){7-7} \cmidrule(lr){8-8} \cmidrule(lr){9-9} \cmidrule(lr){10-10} \cmidrule(lr){11-11} \cmidrule(lr){12-12} \cmidrule(lr){13-13} \cmidrule(lr){14-14} \cmidrule(lr){15-15}
\textsc{unsupervised} & {.211} & {.205} & {.180} & {.003} & {.096} & {.007} & {.166} & {.200} & {.000} & {.000} & {.237} & {.179} & {.000} & {.001} \\
\hdashline
\rowcolor{Gray}
\textsc{5k:orthg-super} & {.251} & {.187} & {.116} & {.195} & {.154} & {.210} & {.155} & {.218} & {.088} & {.085} & {.234} & {.123} & {.056} & {.204} \\
\textsc{5k:orthg+sl+sym} & {.262} & {.205} & {.130} & {.206} & {.165} & {.233} & {.167} & {.245} & {.095} & {.092} & {.255} & {.138} & {.061} & {.221} \\
\rowcolor{Gray}
\textsc{5k:full-super} & {.372} & {.297} & {.243} & {.282} & {.217} & {.292} & {.245} & {.308} & {.169} & {.154} & {.327} & {.214} & {.127} & {.257} \\
\textsc{5k:full+sl} & {.310} & {.233} & {.182} & {.225} & {.133} & {.252} & {.172} & {.258} & {.130} & {.069} & {.248} & {.188} & {.069} & {.203} \\
\rowcolor{Gray}
\textsc{5k:full+sl+nod} & {.337} & {.285} & {.226} & {.257} & {.165} & {.288} & {.216} & {.290} & {.160} & {.117} & {.292} & {.217} & {.114} & {.242} \\
\textsc{5k:full+sl+sym} & {.376} & {.320} & {.265} & {.293} & {.216} & {.318} & {.251} & {.326} & {.172} & {.143} & {.340} & {.253} & {.139} & {.275} \\
\hdashline
\rowcolor{Gray}
\textsc{1k:orthg-super} & {.093} & {.051} & {.036} & {.069} & {.040} & {.049} & {.026} & {.069} & {.024} & {.023} & {.101} & {.026} & {.013} & {.076} \\
\textsc{1k:orthg+sl+sym} & {.166} & {.094} & {.060} & {.133} & {.066} & {.097} & {.064} & {.147} & {.044} & {.040} & {.188} & {.048} & {.014} & {.122} \\
\rowcolor{Gray}
\textsc{1k:full-super} & {.153} & {.088} & {.083} & {.112} & {.068} & {.065} & {.046} & {.103} & {.048} & {.036} & {.138} & {.048} & {.025} & {.091} \\
\textsc{1k:full+sl} & {.234} & {.231} & {.182} & {.225} & {.136} & {.245} & {.175} & {.258} & {.135} & {.077} & {.245} & {.189} & {.060} & {.201} \\
\rowcolor{Gray}
\textsc{1k:full+sl+nod} & {.307} & {.284} & {.215} & {.256} & {.164} & {.286} & {.208} & {.292} & {.159} & {.105} & {.291} & {.220} & {.099} & {.237} \\
\textsc{1k:full+sl+sym} & {.352} & {.305} & {.248} & {.271} & {.172} & {.306} & {.213} & {.308} & {.155} & {.103} & {.317} & {.238} & {.077} & {.255} \\
\bottomrule
\end{tabular}
\end{adjustbox}
}
\vspace{-1.5mm}
\caption{All BLI scores (MRR) with Georgian (\textsc{ka}) as the source language. $5k$ and $1k$ denote the seed dictionary $D_0$ size for (weakly) supervised methods. See Table~\ref{tab:config} for a brief description of each model configuration.}
\label{tab:ka-sup}
\end{table*}

\begin{table*}[t]
\def\arraystretch{0.99}
\centering
{\footnotesize
\begin{adjustbox}{max width=\linewidth}
\begin{tabular}{l cccccccccccccc}
\toprule
{} & \multicolumn{14}{c}{\textbf{Korean}: \textsc{ko}-} \\
\cmidrule(lr){2-15}
{} & {-\textsc{bg}} & {-\textsc{ca}} & {-\textsc{eo}} & {-\textsc{et}} & {-\textsc{eu}} & {-\textsc{fi}} & {-\textsc{he}} & {-\textsc{hu}} & {-\textsc{id}} & {-\textsc{ka}} & {-\textsc{lt}} & {-\textsc{no}} & {-\textsc{th}} & {-\textsc{tr}} \\
\cmidrule(lr){2-2} \cmidrule(lr){3-3} \cmidrule(lr){4-4} \cmidrule(lr){5-5} \cmidrule(lr){6-6} \cmidrule(lr){7-7} \cmidrule(lr){8-8} \cmidrule(lr){9-9} \cmidrule(lr){10-10} \cmidrule(lr){11-11} \cmidrule(lr){12-12} \cmidrule(lr){13-13} \cmidrule(lr){14-14} \cmidrule(lr){15-15}
\textsc{unsupervised} & {.010} & {.001} & {.000} & {.001} & {.001} & {.000} & {.002} & {.000} & {.001} & {.000} & {.000} & {.001} & {.000} & {.000} \\
\hdashline
\rowcolor{Gray}
\textsc{5k:orthg-super} & {.190} & {.183} & {.083} & {.145} & {.102} & {.206} & {.166} & {.238} & {.142} & {.112} & {.156} & {.150} & {.076} & {.213} \\
\textsc{5k:orthg+sl+sym} & {.198} & {.204} & {.088} & {.145} & {.106} & {.229} & {.180} & {.263} & {.162} & {.107} & {.154} & {.166} & {.083} & {.232} \\
\rowcolor{Gray}
\textsc{5k:full-super} & {.289} & {.283} & {.176} & {.242} & {.170} & {.273} & {.257} & {.326} & {.210} & {.178} & {.241} & {.256} & {.174} & {.278} \\
\textsc{5k:full+sl} & {.161} & {.183} & {.065} & {.141} & {.080} & {.180} & {.187} & {.281} & {.162} & {.065} & {.089} & {.193} & {.101} & {.235} \\
\rowcolor{Gray}
\textsc{5k:full+sl+nod} & {.240} & {.262} & {.135} & {.185} & {.114} & {.243} & {.240} & {.320} & {.213} & {.120} & {.138} & {.258} & {.161} & {.270} \\
\textsc{5k:full+sl+sym} & {.303} & {.299} & {.178} & {.229} & {.174} & {.289} & {.282} & {.358} & {.233} & {.176} & {.224} & {.289} & {.176} & {.302} \\
\hdashline
\rowcolor{Gray}
\textsc{1k:orthg-super} & {.052} & {.055} & {.025} & {.031} & {.033} & {.053} & {.026} & {.063} & {.043} & {.024} & {.036} & {.033} & {.014} & {.068} \\
\textsc{1k:orthg+sl+sym} & {.081} & {.071} & {.029} & {.055} & {.039} & {.084} & {.048} & {.122} & {.073} & {.040} & {.055} & {.050} & {.017} & {.112} \\
\rowcolor{Gray}
\textsc{1k:full-super} & {.093} & {.078} & {.045} & {.059} & {.045} & {.066} & {.048} & {.096} & {.060} & {.039} & {.053} & {.047} & {.038} & {.085} \\
\textsc{1k:full+sl} & {.135} & {.187} & {.058} & {.133} & {.073} & {.174} & {.180} & {.278} & {.166} & {.064} & {.098} & {.189} & {.107} & {.233} \\
\rowcolor{Gray}
\textsc{1k:full+sl+nod} & {.227} & {.255} & {.115} & {.179} & {.112} & {.231} & {.223} & {.315} & {.209} & {.111} & {.129} & {.246} & {.134} & {.266} \\
\textsc{1k:full+sl+sym} & {.245} & {.253} & {.110} & {.191} & {.108} & {.266} & {.232} & {.343} & {.206} & {.122} & {.150} & {.244} & {.089} & {.279} \\
\bottomrule
\end{tabular}
\end{adjustbox}
}
\vspace{-1.5mm}
\caption{All BLI scores (MRR) with Korean (\textsc{ko}) as the source language. $5k$ and $1k$ denote the seed dictionary $D_0$ size for (weakly) supervised methods. See Table~\ref{tab:config} for a brief description of each model configuration.}
\label{tab:ko-sup}
\end{table*}

\begin{table*}[t]
\def\arraystretch{0.99}
\centering
{\footnotesize
\begin{adjustbox}{max width=\linewidth}
\begin{tabular}{l cccccccccccccc}
\toprule
{} & \multicolumn{14}{c}{\textbf{Lithuanian}: \textsc{lt}-} \\
\cmidrule(lr){2-15}
{} & {-\textsc{bg}} & {-\textsc{ca}} & {-\textsc{eo}} & {-\textsc{et}} & {-\textsc{eu}} & {-\textsc{fi}} & {-\textsc{he}} & {-\textsc{hu}} & {-\textsc{id}} & {-\textsc{ka}} & {-\textsc{ko}} & {-\textsc{no}} & {-\textsc{th}} & {-\textsc{tr}} \\
\cmidrule(lr){2-2} \cmidrule(lr){3-3} \cmidrule(lr){4-4} \cmidrule(lr){5-5} \cmidrule(lr){6-6} \cmidrule(lr){7-7} \cmidrule(lr){8-8} \cmidrule(lr){9-9} \cmidrule(lr){10-10} \cmidrule(lr){11-11} \cmidrule(lr){12-12} \cmidrule(lr){13-13} \cmidrule(lr){14-14} \cmidrule(lr){15-15}
\textsc{unsupervised} & {.336} & {.194} & {.259} & {.308} & {.001} & {.296} & {.153} & {.286} & {.000} & {.232} & {.000} & {.264} & {.000} & {.181} \\
\hdashline
\rowcolor{Gray}
\textsc{5k:orthg-super} & {.328} & {.224} & {.145} & {.274} & {.165} & {.258} & {.182} & {.288} & {.099} & {.193} & {.093} & {.185} & {.066} & {.220} \\
\textsc{5k:orthg+sl+sym} & {.347} & {.248} & {.164} & {.295} & {.169} & {.290} & {.209} & {.316} & {.108} & {.208} & {.096} & {.207} & {.077} & {.239} \\
\rowcolor{Gray}
\textsc{5k:full-super} & {.462} & {.353} & {.317} & {.394} & {.236} & {.368} & {.299} & {.395} & {.184} & {.284} & {.168} & {.304} & {.162} & {.296} \\
\textsc{5k:full+sl} & {.340} & {.246} & {.264} & {.311} & {.122} & {.273} & {.225} & {.296} & {.136} & {.237} & {.071} & {.258} & {.068} & {.192} \\
\rowcolor{Gray}
\textsc{5k:full+sl+nod} & {.381} & {.323} & {.323} & {.353} & {.161} & {.361} & {.280} & {.361} & {.171} & {.277} & {.115} & {.304} & {.146} & {.245} \\
\textsc{5k:full+sl+sym} & {.437} & {.363} & {.348} & {.383} & {.222} & {.385} & {.316} & {.413} & {.191} & {.304} & {.160} & {.336} & {.168} & {.319} \\
\hdashline
\rowcolor{Gray}
\textsc{1k:orthg-super} & {.160} & {.080} & {.035} & {.110} & {.047} & {.103} & {.040} & {.115} & {.031} & {.065} & {.021} & {.050} & {.018} & {.069} \\
\textsc{1k:orthg+sl+sym} & {.294} & {.171} & {.072} & {.236} & {.075} & {.224} & {.103} & {.216} & {.064} & {.136} & {.038} & {.109} & {.023} & {.135} \\
\rowcolor{Gray}
\textsc{1k:full-super} & {.256} & {.138} & {.102} & {.190} & {.085} & {.143} & {.073} & {.159} & {.058} & {.097} & {.040} & {.081} & {.030} & {.097} \\
\textsc{1k:full+sl} & {.342} & {.250} & {.264} & {.312} & {.124} & {.314} & {.228} & {.294} & {.133} & {.240} & {.065} & {.263} & {.052} & {.197} \\
\rowcolor{Gray}
\textsc{1k:full+sl+nod} & {.384} & {.319} & {.325} & {.353} & {.154} & {.363} & {.280} & {.358} & {.168} & {.277} & {.107} & {.305} & {.126} & {.245} \\
\textsc{1k:full+sl+sym} & {.408} & {.345} & {.332} & {.361} & {.181} & {.380} & {.286} & {.399} & {.168} & {.288} & {.109} & {.322} & {.094} & {.302} \\
\bottomrule
\end{tabular}
\end{adjustbox}
}
\vspace{-1.5mm}
\caption{All BLI scores (MRR) with Lithuanian (\textsc{lt}) as the source language. $5k$ and $1k$ denote the seed dictionary $D_0$ size for (weakly) supervised methods. See Table~\ref{tab:config} for a brief description of each model configuration.}
\label{tab:lt-sup}
\end{table*}

\begin{table*}[t]
\def\arraystretch{0.99}
\centering
{\footnotesize
\begin{adjustbox}{max width=\linewidth}
\begin{tabular}{l cccccccccccccc}
\toprule
{} & \multicolumn{14}{c}{\textbf{Norwegian}: \textsc{no}-} \\
\cmidrule(lr){2-15}
{} & {-\textsc{bg}} & {-\textsc{ca}} & {-\textsc{eo}} & {-\textsc{et}} & {-\textsc{eu}} & {-\textsc{fi}} & {-\textsc{he}} & {-\textsc{hu}} & {-\textsc{id}} & {-\textsc{ka}} & {-\textsc{ko}} & {-\textsc{lt}} & {-\textsc{th}} & {-\textsc{tr}} \\
\cmidrule(lr){2-2} \cmidrule(lr){3-3} \cmidrule(lr){4-4} \cmidrule(lr){5-5} \cmidrule(lr){6-6} \cmidrule(lr){7-7} \cmidrule(lr){8-8} \cmidrule(lr){9-9} \cmidrule(lr){10-10} \cmidrule(lr){11-11} \cmidrule(lr){12-12} \cmidrule(lr){13-13} \cmidrule(lr){14-14} \cmidrule(lr){15-15}
\textsc{unsupervised} & {.326} & {.380} & {.302} & {.296} & {.001} & {.348} & {.256} & {.388} & {.319} & {.207} & {.000} & {.234} & {.000} & {.291} \\
\hdashline
\rowcolor{Gray}
\textsc{5k:orthg-super} & {.316} & {.359} & {.180} & {.284} & {.171} & {.344} & {.224} & {.368} & {.270} & {.189} & {.148} & {.264} & {.101} & {.280} \\
\textsc{5k:orthg+sl+sym} & {.334} & {.390} & {.207} & {.310} & {.189} & {.378} & {.247} & {.401} & {.300} & {.213} & {.168} & {.277} & {.111} & {.301} \\
\rowcolor{Gray}
\textsc{5k:full-super} & {.394} & {.424} & {.323} & {.389} & {.261} & {.396} & {.319} & {.441} & {.306} & {.291} & {.220} & {.366} & {.188} & {.325} \\
\textsc{5k:full+sl} & {.329} & {.390} & {.309} & {.303} & {.200} & {.355} & {.258} & {.389} & {.315} & {.196} & {.141} & {.240} & {.127} & {.292} \\
\rowcolor{Gray}
\textsc{5k:full+sl+nod} & {.374} & {.439} & {.381} & {.345} & {.253} & {.387} & {.292} & {.426} & {.341} & {.235} & {.193} & {.281} & {.182} & {.319} \\
\textsc{5k:full+sl+sym} & {.422} & {.457} & {.377} & {.395} & {.328} & {.419} & {.353} & {.452} & {.340} & {.298} & {.250} & {.351} & {.197} & {.341} \\
\hdashline
\rowcolor{Gray}
\textsc{1k:orthg-super} & {.166} & {.162} & {.069} & {.120} & {.041} & {.153} & {.065} & {.166} & {.121} & {.054} & {.040} & {.110} & {.021} & {.104} \\
\textsc{1k:orthg+sl+sym} & {.284} & {.302} & {.125} & {.233} & {.078} & {.307} & {.136} & {.323} & {.222} & {.111} & {.064} & {.198} & {.028} & {.191} \\
\rowcolor{Gray}
\textsc{1k:full-super} & {.203} & {.198} & {.128} & {.172} & {.075} & {.153} & {.078} & {.206} & {.132} & {.088} & {.057} & {.132} & {.032} & {.123} \\
\textsc{1k:full+sl} & {.329} & {.389} & {.311} & {.303} & {.193} & {.357} & {.258} & {.389} & {.317} & {.195} & {.147} & {.236} & {.125} & {.293} \\
\rowcolor{Gray}
\textsc{1k:full+sl+nod} & {.373} & {.437} & {.374} & {.341} & {.249} & {.389} & {.291} & {.425} & {.338} & {.229} & {.189} & {.281} & {.178} & {.319} \\
\textsc{1k:full+sl+sym} & {.411} & {.444} & {.374} & {.371} & {.300} & {.412} & {.336} & {.443} & {.339} & {.268} & {.228} & {.315} & {.140} & {.332} \\
\bottomrule
\end{tabular}
\end{adjustbox}
}
\vspace{-1.5mm}
\caption{All BLI scores (MRR) with Norwegian (\textsc{no}) as the source language. $5k$ and $1k$ denote the seed dictionary $D_0$ size for (weakly) supervised methods. See Table~\ref{tab:config} for a brief description of each model configuration.}
\label{tab:no-sup}
\end{table*}

\begin{table*}[t]
\def\arraystretch{0.99}
\centering
{\footnotesize
\begin{adjustbox}{max width=\linewidth}
\begin{tabular}{l cccccccccccccc}
\toprule
{} & \multicolumn{14}{c}{\textbf{Thai}: \textsc{th}-} \\
\cmidrule(lr){2-15}
{} & {-\textsc{bg}} & {-\textsc{ca}} & {-\textsc{eo}} & {-\textsc{et}} & {-\textsc{eu}} & {-\textsc{fi}} & {-\textsc{he}} & {-\textsc{hu}} & {-\textsc{id}} & {-\textsc{ka}} & {-\textsc{ko}} & {-\textsc{lt}} & {-\textsc{no}} & {-\textsc{tr}} \\
\cmidrule(lr){2-2} \cmidrule(lr){3-3} \cmidrule(lr){4-4} \cmidrule(lr){5-5} \cmidrule(lr){6-6} \cmidrule(lr){7-7} \cmidrule(lr){8-8} \cmidrule(lr){9-9} \cmidrule(lr){10-10} \cmidrule(lr){11-11} \cmidrule(lr){12-12} \cmidrule(lr){13-13} \cmidrule(lr){14-14} \cmidrule(lr){15-15}
\textsc{unsupervised} & {.000} & {.000} & {.000} & {.000} & {.000} & {.000} & {.000} & {.000} & {.002} & {.000} & {.000} & {.000} & {.000} & {.000} \\
\hdashline
\rowcolor{Gray}
\textsc{5k:orthg-super} & {.151} & {.090} & {.066} & {.119} & {.074} & {.124} & {.123} & {.123} & {.112} & {.086} & {.086} & {.142} & {.115} & {.119} \\
\textsc{5k:orthg+sl+sym} & {.148} & {.096} & {.063} & {.127} & {.073} & {.134} & {.147} & {.139} & {.123} & {.093} & {.090} & {.152} & {.121} & {.130} \\
\rowcolor{Gray}
\textsc{5k:full-super} & {.210} & {.134} & {.087} & {.186} & {.094} & {.173} & {.173} & {.178} & {.141} & {.116} & {.112} & {.214} & {.162} & {.177} \\
\textsc{5k:full+sl} & {.051} & {.058} & {.017} & {.071} & {.035} & {.041} & {.122} & {.051} & {.119} & {.018} & {.036} & {.044} & {.050} & {.071} \\
\rowcolor{Gray}
\textsc{5k:full+sl+nod} & {.101} & {.098} & {.041} & {.129} & {.062} & {.123} & {.186} & {.097} & {.167} & {.061} & {.069} & {.140} & {.093} & {.120} \\
\textsc{5k:full+sl+sym} & {.174} & {.123} & {.073} & {.164} & {.093} & {.167} & {.203} & {.160} & {.170} & {.126} & {.097} & {.215} & {.147} & {.160} \\
\hdashline
\rowcolor{Gray}
\textsc{1k:orthg-super} & {.042} & {.021} & {.018} & {.046} & {.022} & {.030} & {.037} & {.035} & {.045} & {.028} & {.031} & {.057} & {.027} & {.041} \\
\textsc{1k:orthg+sl+sym} & {.066} & {.027} & {.024} & {.062} & {.024} & {.038} & {.062} & {.060} & {.067} & {.045} & {.047} & {.085} & {.050} & {.067} \\
\rowcolor{Gray}
\textsc{1k:full-super} & {.049} & {.027} & {.021} & {.070} & {.029} & {.032} & {.057} & {.044} & {.044} & {.034} & {.040} & {.084} & {.029} & {.052} \\
\textsc{1k:full+sl} & {.049} & {.047} & {.016} & {.059} & {.030} & {.045} & {.112} & {.054} & {.122} & {.018} & {.037} & {.047} & {.057} & {.067} \\
\rowcolor{Gray}
\textsc{1k:full+sl+nod} & {.086} & {.083} & {.036} & {.121} & {.046} & {.104} & {.174} & {.096} & {.162} & {.047} & {.063} & {.119} & {.094} & {.116} \\
\textsc{1k:full+sl+sym} & {.108} & {.084} & {.036} & {.128} & {.057} & {.094} & {.152} & {.111} & {.168} & {.073} & {.065} & {.145} & {.098} & {.121} \\
\bottomrule
\end{tabular}
\end{adjustbox}
}
\vspace{-1.5mm}
\caption{All BLI scores (MRR) with Thai (\textsc{th}) as the source language. $5k$ and $1k$ denote the seed dictionary $D_0$ size for (weakly) supervised methods. See Table~\ref{tab:config} for a brief description of each model configuration.}
\label{tab:th-sup}
\end{table*}

\begin{table*}[t]
\def\arraystretch{0.99}
\centering
{\footnotesize
\begin{adjustbox}{max width=\linewidth}
\begin{tabular}{l cccccccccccccc}
\toprule
{} & \multicolumn{14}{c}{\textbf{Turkish}: \textsc{tr}-} \\
\cmidrule(lr){2-15}
{} & {-\textsc{bg}} & {-\textsc{ca}} & {-\textsc{eo}} & {-\textsc{et}} & {-\textsc{eu}} & {-\textsc{fi}} & {-\textsc{he}} & {-\textsc{hu}} & {-\textsc{id}} & {-\textsc{ka}} & {-\textsc{ko}} & {-\textsc{lt}} & {-\textsc{no}} & {-\textsc{th}} \\
\cmidrule(lr){2-2} \cmidrule(lr){3-3} \cmidrule(lr){4-4} \cmidrule(lr){5-5} \cmidrule(lr){6-6} \cmidrule(lr){7-7} \cmidrule(lr){8-8} \cmidrule(lr){9-9} \cmidrule(lr){10-10} \cmidrule(lr){11-11} \cmidrule(lr){12-12} \cmidrule(lr){13-13} \cmidrule(lr){14-14} \cmidrule(lr){15-15}
\textsc{unsupervised} & {.214} & {.276} & {.000} & {.001} & {.001} & {.192} & {.208} & {.332} & {.307} & {.000} & {.000} & {.073} & {.260} & {.000} \\
\hdashline
\rowcolor{Gray}
\textsc{5k:orthg-super} & {.265} & {.288} & {.112} & {.212} & {.164} & {.247} & {.214} & {.328} & {.267} & {.131} & {.126} & {.209} & {.226} & {.103} \\
\textsc{5k:orthg+sl+sym} & {.280} & {.314} & {.121} & {.229} & {.172} & {.273} & {.241} & {.357} & {.288} & {.146} & {.145} & {.217} & {.261} & {.121} \\
\rowcolor{Gray}
\textsc{5k:full-super} & {.344} & {.360} & {.215} & {.307} & {.230} & {.294} & {.319} & {.378} & {.336} & {.205} & {.196} & {.295} & {.311} & {.170} \\
\textsc{5k:full+sl} & {.236} & {.300} & {.146} & {.172} & {.155} & {.249} & {.230} & {.338} & {.308} & {.113} & {.140} & {.089} & {.274} & {.112} \\
\rowcolor{Gray}
\textsc{5k:full+sl+nod} & {.283} & {.355} & {.221} & {.216} & {.198} & {.281} & {.275} & {.375} & {.358} & {.168} & {.171} & {.137} & {.325} & {.176} \\
\textsc{5k:full+sl+sym} & {.351} & {.376} & {.238} & {.309} & {.244} & {.322} & {.323} & {.397} & {.370} & {.229} & {.214} & {.280} & {.346} & {.183} \\
\hdashline
\rowcolor{Gray}
\textsc{1k:orthg-super} & {.107} & {.103} & {.024} & {.065} & {.048} & {.089} & {.058} & {.143} & {.112} & {.029} & {.036} & {.059} & {.058} & {.023} \\
\textsc{1k:orthg+sl+sym} & {.198} & {.184} & {.039} & {.130} & {.076} & {.160} & {.122} & {.303} & {.203} & {.055} & {.063} & {.112} & {.121} & {.040} \\
\rowcolor{Gray}
\textsc{1k:full-super} & {.150} & {.133} & {.052} & {.112} & {.062} & {.093} & {.076} & {.167} & {.131} & {.053} & {.050} & {.099} & {.073} & {.028} \\
\textsc{1k:full+sl} & {.230} & {.309} & {.156} & {.167} & {.147} & {.250} & {.223} & {.336} & {.312} & {.108} & {.139} & {.093} & {.273} & {.113} \\
\rowcolor{Gray}
\textsc{1k:full+sl+nod} & {.280} & {.354} & {.213} & {.218} & {.183} & {.281} & {.269} & {.371} & {.352} & {.160} & {.167} & {.138} & {.323} & {.166} \\
\textsc{1k:full+sl+sym} & {.327} & {.364} & {.204} & {.274} & {.209} & {.310} & {.301} & {.398} & {.363} & {.201} & {.194} & {.215} & {.344} & {.137} \\
\bottomrule
\end{tabular}
\end{adjustbox}
}
\vspace{-1.5mm}
\caption{All BLI scores (MRR) with Turkish (\textsc{tr}) as the source language. $5k$ and $1k$ denote the seed dictionary $D_0$ size for (weakly) supervised methods. See Table~\ref{tab:config} for a brief description of each model configuration.}
\label{tab:tr-sup}
\end{table*}


\end{document}